\documentclass{article}
\PassOptionsToPackage{table,xcdraw}{xcolor}
\usepackage[preprint]{tmlr}
\usepackage{subfigure}
\usepackage{graphicx} % Required for inserting images
\usepackage{wrapfig}  % for wrapping figures in text
\usepackage{tabularx}
\usepackage{booktabs}
\usepackage{natbib}
\usepackage{colortbl}
\usepackage{placeins}
\usepackage{adjustbox}
\usepackage{makecell}
\usepackage{hyperref}
\usepackage{pifont}
\usepackage{cleveref}
\usepackage{listings}  % for code snipets
\crefname{figure}{Figure}{Figures}
\Crefname{figure}{Figure}{Figures}
\crefname{table}{Table}{Tables}
\Crefname{table}{Table}{Tables}
\crefname{section}{Section}{Sections}
\Crefname{section}{Section}{Sections}
\crefname{appendix}{Appendix}{Appendices}
\Crefname{appendix}{Appendix}{Appendices}

\usepackage[scaled=0.80]{roboto-mono}         % Better monospace font
\usepackage{caption}                                % Better control over caption
\newcommand{\trafficlight}[1]{%
  \fboxsep=1pt%
  \fboxrule=0.5pt%
  \newcommand{\lightcircle}[2]{%
    \fcolorbox{black}{##2}{\textcolor{##2}{\rule{0.8em}{0.8em}}}%
  }%
  \ifnum#1=1\lightcircle{black}{red}\else\lightcircle{red!40}{red!20}\fi%
  \ifnum#1=2\lightcircle{black}{yellow}\else\lightcircle{yellow!40}{yellow!20}\fi%
  \ifnum#1=3\lightcircle{black}{green}\else\lightcircle{green!40}{green!20}\fi%
}

\title{TapeAgents: a Holistic Framework for Agent Development and Optimization}

\date{\today}

\begin{document}

\maketitle

\renewcommand{\thempfootnote}{\arabic{mpfootnote}}

\vspace{-1.2cm}
Core contributors\footnote{Authors are in alphabetical order.}:\\
\textbf{Dzmitry~Bahdanau}\textsuperscript{†,*}
\quad\textbf{Nicolas~Gontier}\textsuperscript{†}
\quad\textbf{Gabriel~Huang}\textsuperscript{†}\\
\quad\textbf{Ehsan~Kamalloo}\textsuperscript{†}
\quad\textbf{Rafael~Pardinas}\textsuperscript{†}
\quad\textbf{Alex~Piché}\textsuperscript{†}\\
\quad\textbf{Torsten~Scholak}\textsuperscript{†}
\quad\textbf{Oleh~Shliazhko}\textsuperscript{†,*}
\quad\textbf{Jordan~Prince~Tremblay}\textsuperscript{†,*}

Contributors\footnotemark[1]:\\
\textbf{Karam~Ghanem}\footnote{Work performed while at \textit{ServiceNow}}
\quad \textbf{Soham~Parikh}\textsuperscript{‡}
\quad \textbf{Mitul~Tiwari}\footnotemark[2]
\quad \textbf{Quaizar~Vohra}\textsuperscript{‡}

% Affiliation
% {Core Contributor, † Contributor. Authors are in alphabetical order.
\textsuperscript{†}\textit{ServiceNow Research}
\quad \textsuperscript{‡}\textit{ServiceNow}

Corresponding authors (*): \textit{\{dzmitry.bahdanau,oleh.shliazhko,jordanprince.t\}@servicenow.com}

\vspace{0.5cm}

% \begin{center}
% \vspace*{-10mm}
% Manuscript version: \today
% \end{center}

\begin{abstract}
We present TapeAgents,\footnote{\href{https://github.com/ServiceNow/TapeAgents}{https://github.com/ServiceNow/TapeAgents}} an agent framework built around a granular, structured log (\textbf{tape}) of the agent session that also plays the role of the session's resumable state. In TapeAgents we leverage tapes  to facilitate all stages of the LLM Agent development lifecycle. 
The agent reasons by processing the tape and the LLM output to produce new thought and action steps and append them to the tape. The environment then reacts to the agent's actions by likewise appending observation steps to the tape. 
% DIMA: old version, with the word "replayable" 
% We present TapeAgents,\footnote{\href{https://github.com/ServiceNow/TapeAgents}{https://github.com/ServiceNow/TapeAgents}} an agent framework that leverages a structured, replayable log (\textbf{tape}) of the agent session to facilitate all stages of the LLM Agent development lifecycle. 
% 
% In TapeAgents, the agent reasons by processing the tape and the LLM output to produce new thought and action steps and append them to the tape. The environment then reacts to the agent's actions by likewise appending observation steps to the tape. 
% 
By virtue of this tape-centred design, TapeAgents can provide AI practitioners with holistic end-to-end support. At the development stage, tapes facilitate session persistence, agent auditing, and step-by-step debugging. Post-deployment, one can reuse tapes for evaluation, fine-tuning, and prompt-tuning; crucially, one can adapt tapes from other agents or use revised historical tapes. 
In this report, we explain the TapeAgents design in detail. We demonstrate possible applications of TapeAgents with several concrete examples of building monolithic agents and multi-agent teams, of optimizing agent prompts and finetuning the agent's LLM. We present tooling prototypes and report a case study where we use TapeAgents to finetune a Llama-3.1-8B form-filling assistant to perform as well as GPT-4o while being orders of magnitude cheaper.  
% DIMA: older version
% We compare TapeAgents to prior work and find that our framework's friendliness to both development and data-driven optimization comes from building LLM Agents as resumable modular state machines with structured configuration that make granular structured logs~---~a trait combination that prior frameworks do not possess.
Lastly, our comparative analysis shows that TapeAgents's advantages over prior frameworks stem from our novel design of the LLM agent 
as a resumable, modular state machine with a structured configuration, that generates granular, structured logs and that can transform these logs into training text~---~a unique combination of features absent in previous work.

%% FROM CHATGPT
% Our comparative analysis demonstrates that TapeAgents not only streamlines development but also significantly enhances data-driven optimization---advantages unmatched by prior frameworks. This superiority arises from our novel design of LLM Agents as resumable, modular state machines with structured configurations that generate granular, structured logs---a unique combination of features absent in previous work.

% We compare TapeAgents to prior agent frameworks and find that TapeAgents is the first one that helps the practitioner to build, debug, serve, and optimize their agent.%
\hfill
\mbox{\textsl{Manuscript version: \today}}

% We compare TapeAgents to prior work and find that its friendliness to both developers and applied scientists comes from the agent being a resumable modular state machine with a structured config that makes granular structured logs~---~a traint combination that prior frameworks do not possess.
\end{abstract}

\FloatBarrier

\section{Introduction}
In the coming years, we will likely witness widespread deployments of Large Language Model (LLM) Agents: complex user-facing and background workflows that interleave traditional programming with LLM-based intelligence. This big paradigm shift in software architecture will greatly challenge AI practitioners who put LLM agents to work. The agent developers and applied scientists will have to troubleshoot and improve systems that operate in non-stationary environments and deal with non-deterministic LLM behavior and the LLM's often fragile instruction following. For the LLM agent adoption to go smoothly and lead to good outcomes, it is crucial that agent developers and applied scientists operate in appropriate frameworks that enable effective tooling. Developers and researchers have recently proposed many agentic frameworks that support practitioners at different stages of the agent development lifecycle. Several frameworks, like LangChain~\citep{Chase_LangChain_2022}, CrewAI and AutoGen~\citep{wu2023autogen}, help developers quickly build an agent using low-code paradigms, such as prompt-chaining or multi-agent teams. Others, like LangGraph~\citep{Chase_LangGraph_2023}, offer low-level support in achieving resumability, asynchronous execution, concurrency and instrumentation. At the other end of the spectrum are frameworks built by researchers like DSPy~\citep{khattab2023dspy}, TextGrad~\citep{yuksekgonul2024textgradautomaticdifferentiationtext} and Agents~\citep{zhou2023agents}, that usually focus on data-driven optimization of the agent performance with model finetuning and prompt-tuning algorithms, while putting less emphasis on the needs of the agent developers.

\begin{figure}[t]
  \centering
  \includegraphics[width=1\linewidth,clip]{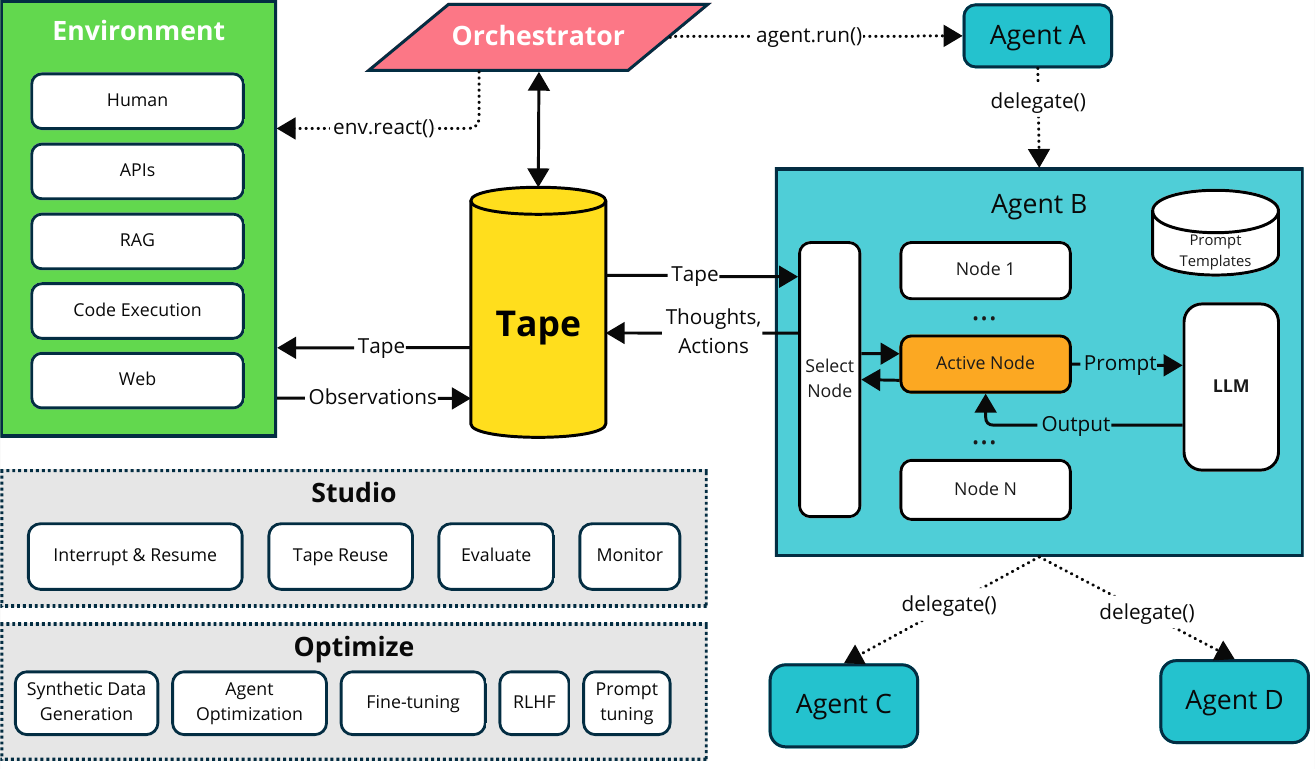}
  \caption{\textbf{TapeAgents at a glance}. The orchestrator alternates between running the agent and the environment interacting with each other via adding steps to the tape: a comprehensive, replayable semantic-level log. Agents are composed of basic reasoning units that we call nodes.
  The agents are organized in hierarchical teams, with one agent being active at a time.
  % The core idea is that the interactions between the agent, the LLM and the environment are mediated by orchestrator and go through the tape. 
  The tape and the agent configurations are highly structured and linked with rich metadata that supports the implementation of broadly usable developer tools (collectively called \textit{Studio}) and optimization methods (collectively called \textit{Optimize}).
  % for Agentic System. Every Agent step, Environment observation and LLM output is recorded on a \textit{Tape}. TapeAgents \textit{Studio} helps AI Admin to monitor, debug and evaluate LLM Agents. TapeAgents \textit{Optimize} enables AI Admin to generate synthetic data, improve Agents and LLM.
  % Multiple agents are configured in a tree structure, with each agent having its own set of nodes.
  }
  \label{fig:overview}
\end{figure}

In this technical report, we present \textbf{TapeAgents}~---~a new holistic agent framework that supports practitioners at both the agent development and data-driven agent optimization stages. We achieve both objectives by building the framework around a comprehensive, structured, granular, semantic-level log of the agent session that we call a \textbf{tape}, a term that also gives the framework its name (see Figure \ref{fig:overview} for an illustration).
The agents in TapeAgents read the tape to make the LLM prompt and then process the LLM output to append new \textbf{steps} to the tape: \textbf{thought} steps to express reasoning and \textbf{action} steps to request external inputs.
The \textbf{environment} responds to the action steps at the end of the tape with \textbf{observation} steps that it likewise appends to the tape.
The \textbf{orchestrator} invokes the agent and the environment in an alternate fashion and maintains full control over their interactions. By design, the orchestrator can resume from any intermediate tape, which enables session persistence and step-by-step debugging, both key developer requirements for an agent framework. 
For data-driven algorithms, tapes record the attribution of each step to the respective part of the agent configuration, which facilitates training, data generation and automatic prompt-tuning.
Crucially, for both manual debugging and algorithms, agents can reuse lightly adapted tapes from other agents and revise their own tapes. This allows practitioners to maximally benefit from imperfect historical tapes by earlier versions of the agent, both for evaluating the newer versions and for improving them algorithmically. Last but not least, agents stream their intermediate events to the orchestrator to enable delightful interactive experiences.

We invite the reader to start their TapeAgents journey with the technical presentation of the framework in \cref{sec:tapeagents_foundations}. There, we cover the details of agent architecture, agent-environment orchestration, tape content and structure. \Cref{sec:tapeagents_tooling} describes three low-code agent-building framework prototypes on top of TapeAgents: one for monolithic agents, another for multi-agent teams and the third one with easy-to-tune function-like prompts. 
The same section also covers early versions of our Studio toolsuite for development and debugging and our Optimize toolsuite for agent optimization. In \cref{sec:examples}, we present diverse examples of building and optimizing agents using TapeAgents framework and tooling. \Cref{sec:ghreatintro} presents a deeper case study of a key practical TapeAgents use case: optimizing the quality of a cost-effective conversational assistant using tapes from an expensive multi-node Teacher agent. After presenting the framework and the examples we offer the reader a detailed comparison of TapeAgents with prior work in \cref{sec:related_work}. Lastly, \cref{sec:future_work} discusses possible extensions and applications of TapeAgents.
 
% TapeAgents is an experimental  framework to build, debug, serve and optimize AI agents. The key concept of the framework is the Tape: a complete semantic-level log of the agent's session. All Agent-Environment interactions are mediated by the orchestrator and must go through the tape

% In this paper, we describe how to build a tape agent and we argue that TapeAgents and their tapes are a great operating system for agent optimization algorithms. We compare with other frameworks.

% \subsection{Case Studies}

% We demonstrate the advantages and the flexibility of TapeAgents framework with the following 5 examples:
% \begin{itemize}
%     \item We show how a simple TapeAgent can achieve a high performance of X\% on the challenging GAIA benchmark.
%     \item We demonstrate a multimodal TapeAgent that can browse the web.  
%     \item We show how one can build AutoGen-like multi-agent teams using TapeAgents.
%     \item We describe how one can implement DSPy-style prompt-tuning using TapeAgents.
%     \item We show how one can train a conversational TapeAgent to provide a high-quality conversational form-filling experience at low price. First, we build a slow and expensive agent that serves as the gold standard and as a teacher. We experiment with direct distillation of the teacher in a small student (Section~\ref{sec:ghreatintro}).  
%\end{itemize}

\section{TapeAgents: foundations}
\label{sec:tapeagents_foundations}

Our TapeAgents framework proposes an agent-building paradigm that facilitates all stages of the AI Agent development lifecycle. This section presents the framework in a detailed bottom-up approach. First, we introduce the building blocks: the nodes, the agents, and the environment. Then, we explore how these parts can be composed and orchestrated to build a tape-centered system. In this section, we also describe the tape structure and metadata.

\subsection{Nodes and Steps}
As outlined in \cref{fig:overview}, in TapeAgents, one builds the agent from \textbf{nodes}: the basic atoms of intelligence. A node describes one LLM call and the classical symbolic processing of the call's output. The agent will dynamically determine which node to run next based on the tape. Nodes generate new tape entries that we call \textbf{steps}: basic atoms of the agent's memory. Examples of what an agent can do in a step include making a long-term plan, reasoning about how to fulfill the plan or how to use a tool, requesting a tool call. Among these examples, the last one is an \textbf{action} step as it requests interaction with or has an impact on the agent's environment. The first three examples are \textbf{thoughts}: the agent's inner reasoning steps. The remaining step type in TapeAgents is the \textbf{observations} that the agent receives from its environment in response to the agent's actions. The reader can find an example tape with color-coded actions, thoughts and observations in \Cref{fig:view_agent_run}. In TapeAgents we often define a tape type by specifying what specific actions, thoughts, and observations classes it can contain, though all such tapes are currently merely aliases for the one and only \texttt{tapeagents.core.Tape} class.

% (Nicolas, edited by Dima)
A typical node uses an LLM to generate tape steps. One defines this process with two node methods: \texttt{make\_prompt} and \texttt{generate\_steps}.
First, the node constructs the LLM prompt through its \texttt{make\_prompt} method that has the following Python signature:
% (Dima) 
% A typical node generates tape steps based on the stream of tokens that the agent receives from the last LLM call. To trigger the LLM call, the node constructs the LLM prompt through its \texttt{make\_prompt} method that has the following Python signature:
\begin{equation}
\texttt{def make\_prompt(self, agent, tape) -> Prompt}
\end{equation}
Some nodes perform only the conventional non-neural computation, like taking a branching decision. These nodes can use the default \texttt{make\_prompt} implementation that produces a null prompt. Note that the node does not call the LLM directly but only makes a prompt. This is a deliberate design decision to keep all node methods pure functions, i.e. deterministic functions with no side effects.

Second, the node generates steps based on the stream of tokens that it receives from the last LLM call.
% By default, the agent will be calling its nodes sequentially and append the steps they create to the tape that the agent is asked to continue.
One defines the step-generating behavior of a new node class in its \texttt{generate\_steps} method:
\begin{equation}
\texttt{def generate\_steps(self, agent, tape, llm\_stream) -> Generator[Step | PartialStep]}
\end{equation}
If a node produced a null prompt in its \texttt{make\_prompt} method above, the \texttt{llm\_stream} will also be null.
All nodes must generate \texttt{Step} objects; some can also parse the LLM token stream incrementally to produce partial steps which the agent will pass through to the application without adding them to the tape. \Cref{fig:agent_run} shows how the agent runs one node and adds the resulting steps to the tape, along with the relationship between \verb|make_prompt| and \verb|generate_steps|. 

By default, the agent calls its nodes sequentially and appends the steps they create to the tape that the agent is asked to continue.
A code example of a node implementing these two methods can be seen in Appendix~\ref{appendix:multi_agent_code} under \texttt{class SearchAgentMainNode}.
% DIMA: v0 version

% By default, the agent will be calling its nodes sequentially and append the steps they create to its \textbf{tape}. To define the step-generating behavior of a new node class, one must implement its \texttt{generate\_steps} method with the following Python signature:

%\begin{equation}
%\texttt{def generate\_steps(self, agent, tape, llm\_stream) -> Generator[Step | PartialStep]}
%\end{equation}
% All nodes must generate \texttt{Step} objects; some can parse the LLM token stream incrementally to produce partial steps which the agent will pass through to the application without adding them to the tape. 

% To receive a non-null stream of tokens from the LLM, the node must make the LLM prompt with the respective method:
% \begin{equation}
% \texttt{def make\_prompt(self, agent, tape) -> Prompt}
% \end{equation}
% Note that some nodes perform only the conventional non-neural computation, like taking a branching decision. These nodes can use the default \texttt{make\_prompt} implementation that produces a null prompt. Note that the node does not call the LLM directly but only makes a prompt for the agent's internal orchestration code that calls the LLM. This is a deliberate design decision to keep all node methods pure functions, i.e. deterministic functions with no side effects. 
% %
% Figure \ref{fig:agent_run} shows how the agent runs one node and adds the resulting steps to the tape. 

\subsubsection{Nodes That Can Make Training Data}
\label{sec:make_llm_output}

Some nodes also implement the reverse direction~---~make the LLM output that would be required to produce the steps at a given index in the tape. The respective node method is
\begin{equation}
\texttt{def make\_llm\_output(self, agent, tape, index) -> LLMOutput}
\end{equation}
This method is crucial for making fine-tuning data.

\subsection{Agents}

\begin{wrapfigure}[24]{R}{0.4\textwidth} % [#] = number of lines to cut to make space for the figure
\vspace{-.33in}  % to move the figure up a little more than the default position
  \centering
  \includegraphics[width=0.4\textwidth]{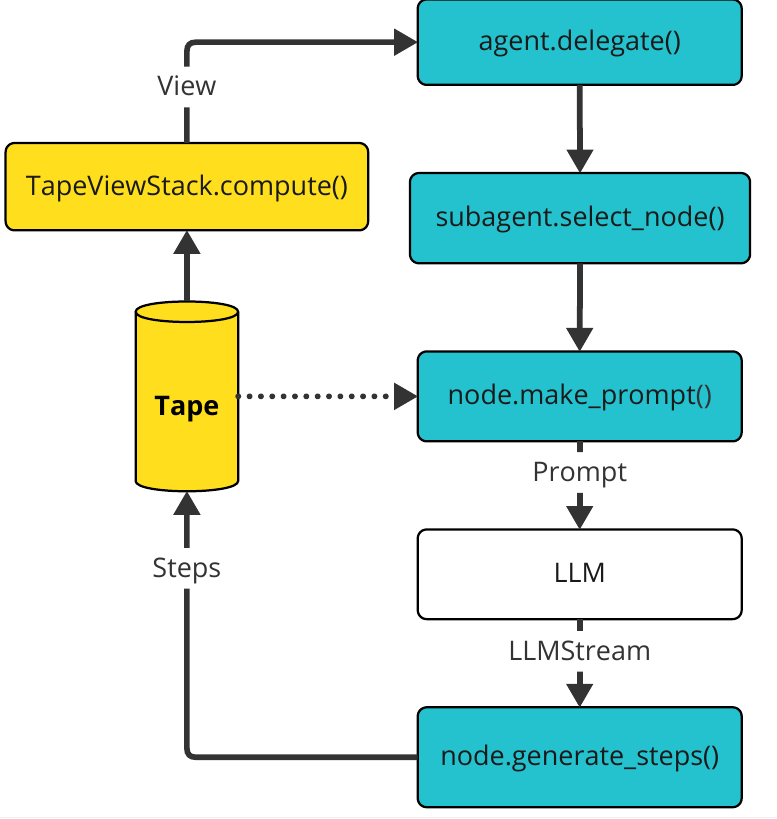}
  \caption{A reasoning loop of an agent in TapeAgents. The root agent delegates to a subagent, the subagent selects the node, the node makes the prompt. The subagent calls the LLM with the prompt and lets the node process the resulting stream of tokens (LLMStream) that the root agent will then append to the tape.}
  \label{fig:agent_run}
\end{wrapfigure}

Like nodes, a TapeAgent agent generates steps and makes a new tape by appending the generated steps to the input tape. Specifically, \texttt{agent.run(tape)} runs an iterative reasoning loop that, at every iteration, selects a node, lets it make the prompt and generates the next steps (see~\cref{fig:agent_run}). By default, the agent will run its nodes sequentially, unless the previous node produced a special \texttt{SetNextNode} step that explicitly the determines the next that will run next. The loop continues as long as the nodes only generate thoughts. When a node produces an action, the agent stops and returns a new tape with the generated steps from all iterations appended to it. More precisely, \texttt{agent.run(tape)} returns an \texttt{AgentStream} object for streaming events like partial tapes and steps, but the final new agent tape is easy to extract from the stream object using \texttt{AgentStream.get\_final\_tape()} method.
 
An agent may have \textbf{subagents} for whom this agent is the \textbf{manager}. The subagents can have further subagents, which gives rise to a hierarchical agent organization with a single manager-free root agent on top. Given an input tape, the root agent determines the next active organization member to which it will delegate the generation of next steps. By default, the root agent makes the delegation decision by looking at the special \texttt{Call} and \texttt{Respond} thoughts. When an agent A wants the root to delegate to an agent B, A will append \texttt{Call(agent\_name="B", content=...)} thought to the tape with an optional free-form message in the \texttt{content} field. When B responds by appending \texttt{Respond(content=...)}, A becomes active again. Note that both \texttt{Call} and \texttt{Respond} will affect the delegation logic \textit{at the next agent iteration}. To sum up the delegation description, the root delegates to the agent that was called last and has not responded yet.
See \cref{fig:view_agent_run} for an example of communication between a financial analyst agent and its web search helper. See Appendix~\ref{appendix:multi_agent_code} for a listing of the complete code for this example.

\subsubsection{Tape Views}

In a multi-agent system different agents usually are expected to maintain their  respective different states. In TapeAgents the tape combines the states of all agents in the team. To determine what they should base their acting and reasoning on, most agents use \textbf{tape view stack}. For each agent that has not responded yet, the \textbf{view} contains the steps that this agent can see. Specifically, for an agent A the view contains the tape's steps starting from the \texttt{Call} step that initiated A's activity and excluding the inner steps of the subagents that A called (see Figure \ref{fig:view_agent_run}). Note that the default tape view stack is only one possible way to determine what parts of the tape each agent should see. For example, one can let an agent see all its prior \texttt{Call} steps to enable an inter-agent conversation history.

% In many cases, a middle-level agent A will make its prompts only using a subset of the tape's steps starting from A's last \texttt{Call} message and excluding the inner steps of the subagents that A called.
% To keep track of the steps each subagent can see, and to select the active agent and node, most agents compute the \textbf{tape view stack} from the tape and delegate to the agent whose view is at the top. For each agent that has not responded yet, the \textbf{view} contains the steps that this agent can see and the next node that the agent should run when it becomes active (see Figure \ref{fig:view_agent_run}).
%To track the next node, the \texttt{TapeViewStack.compute} function increments the top view's next node pointer when it sees (from step metadata) that the current node has started running (see more on step metadata in Section \ref{sec:metadata}). Exceptionally when \texttt{TapeViewStack.compute} encounters a special \texttt{SetNextNode(next\_node=...)} thought, it changes the next node pointer to the value that the thought carries. A common use-case for \texttt{SetNextNode} is to implement looping within an agent. 

% DIMA: is below kinda clear from the context?
%method incrementally processes the steps' metadata, the special agent communication steps %(\texttt{Call} and \texttt{Respond}), as well as the control flow steps (\texttt{Pass} and \texttt{SetNextNode}) that explicitly control the choice of the node at the next iteration.

A reader familiar with how Python interpreter works can find agents similar to Python functions, node similar to lines of Python code, steps similar to Python bytecode instructions, the tape view stack similar to the Python call stack and tape views similar to Python frames.

\subsubsection{Optimizable Agents}
Agent optimization algorithms tune agent prompts \citep{khattab2023dspy,pryzant-etal-2023-automatic,zhou2023large} or alter agent structure \citep{hu2024automateddesignagenticsystems} in order to maximize the agent's performance. 
% DIMA: here we mix aligment papers with agent structure papers
%\citep{bai2022constitutionalaiharmlessnessai,shinn2024reflexion,yuan2024selfrewarding,wu2024meta}
To make such algorithms applicable to as many agents as possible, we standardize the structure of the agent configuration. We achieve this by making \texttt{tapeagents.agent.Agent} a Pydantic model\footnote{Pydantic models, from the Python \texttt{pydantic} package, provide robust data validation and parsing, ensuring that input data is properly structured and typed. This not only improves reliability by catching errors early but also simplifies code by automatically handling serialization, deserialization, and type conversions.} with the following mandatory fields: \texttt{.llms} for the LLM configurations, \texttt{.templates} for the prompt templates, \texttt{.nodes} for the nodes,  and \texttt{.subagents} for the subagents.

% DIMA: old version, better one above.
% To support agent optimization algorithms including prompt-tuning, we enforce a universal agent configuration schema. An agent must  At the implementation level, this is achieved by defining all agent and node classes as Pydantic models.

Agents can also make training data for the LLM that they use. An agent's \verb|agent.make_training_text(tape)| method reconstructs the LLM calls from a given tape, validates the reconstruction by replaying the step generation and returns training text characters. Internally, \texttt{agent.make\_training\_text} uses \texttt{node.make\_llm\_output} method introduced in \cref{sec:make_llm_output}; hence all nodes must implement this method for the agent to be trainable.

\subsection{Environment}
Just like nodes and agents, the environment in TapeAgents makes a new tape by adding steps to an existing tape. The main method of an environment object is:
\begin{equation}
\texttt{def react(self, tape) -> Tape}.
\end{equation}
The \texttt{environment.react} searches for the unfulfilled actions in the tape and adds the corresponding \textbf{observation} steps to the tape.
Unlike nodes and agents, the environment may be non-deterministic and have side effects. We encourage agent developers to put all the deterministic and pure-function aspects of the system in the agent part, isolating only non-deterministic, computationally heavy or transactional aspects in the environment part.

\subsection{Orchestration}
To run a TapeAgent-based agentic application, one must alternate between running the root agent (which handles the delegation internally) and calling the environment to react to the agent's actions (see \cref{fig:overview}). While we provide a default \texttt{tapeagents.orchestrator.main\_loop} orchestrator for this purpose, we expect many application developers to build their custom orchestrators to closely control the agent-environment communication and ensure safety or enhance iteration logic.

\subsubsection{Resumption and Replay}

We designed TapeAgents with resumption and replay as key priorities. To resume, one can just restart the orchestration from an intermediate tape. For testing purposes, one can run an agent with replayed observations and LLM outputs and verify that this process leads to the same tape or print the diff otherwise. We found the replay tests to be incredibly helpful in our development work. When applicable, one can also replay the tape's observations (or even some of the agent's steps) in a new session to evaluate a new agent, though the old observations can be implausible if the new tape deviates too much from the old one. 

\subsection{Tape Metadata and LLM Call Database}
\label{sec:metadata}
Regardless of the orchestration method, the implementations of \texttt{agent.run()} and \texttt{environment.react()} ensure that the tape and its steps contain rich metadata, including these fields:
\begin{itemize}
\item \texttt{tape.metadata.author}: 
% v1 by Dima: which agent or environment made this tape by adding steps to another one OR which entity authored this tape from scratch / via revision.
% v2 by Nicolas: which agent or environment made this tape; either by adding steps to it, or by authoring it from scratch.
% v3 by Dima: 
which agent or environment made this tape; either by authoring it, or by adding steps to it, or by making a revision of another tape.
\item \texttt{tape.metadata.parent\_id}:
% v1 by Dima, the id of the tape before the last \texttt{agent.run()}  or environment \texttt{environment.metadata.react()}, if applicable.
% v2 by Nicolas, the ID of the (parent) tape before it was modified by the current agent or environment, if applicable.
% v3 by Dima
the ID of the parent tape of which the current tape is a continuation (when applicable).
\item \texttt{step.metadata.agent}: the hierarchical name of the agent that generated the step.
\item \texttt{step.metadata.node}: the name of the node that generated the step.
\item \texttt{step.metadata.prompt\_id}: % the id of the prompt used that was used in the node run that generated the current step (also serves as the unique id of a node run).
the identifier (id) of the prompt that led to the generation of this step, see the explanation below.
\end{itemize}
When an agent runs a node, the node generates a unique ID for the prompt that it builds at this iteration. The prompt ID thus serves as the unique identifier of a node execution, i.e., of a specific iteration when the node was active. The ID also links the step to the LLM call from the node run so we can trace the origin of each step down to the specific prompt and LLM output. We store the prompt and the output for all LLM calls in an SQLite database. One can view LLM calls as an effective part of the tape in that they are always easily accessible; we don't include them in the tape to keep the latter lightweight.

\begin{figure}[h!]
  \centering
  \begin{minipage}[t]{0.15\textwidth}
    \vspace{0pt}  % Ensures top alignment
    \includegraphics[width=\linewidth, clip]{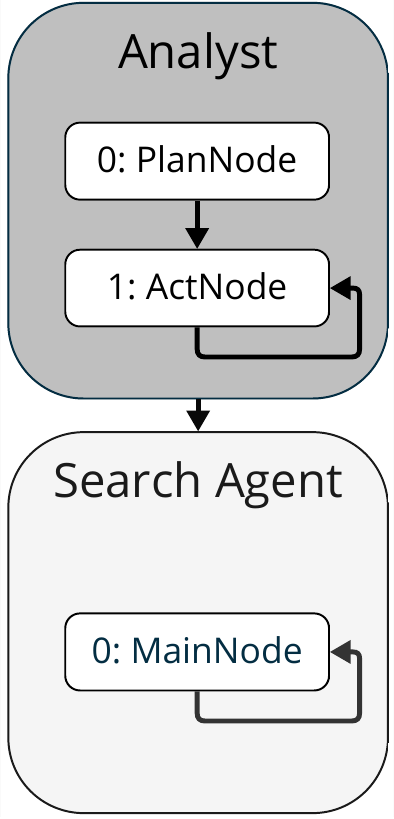}
  \end{minipage}%
  \hspace{10pt}
  \begin{minipage}[t]{0.62\textwidth}
    \vspace{0pt}  % Ensures top alignment
    \includegraphics[width=\linewidth, clip]{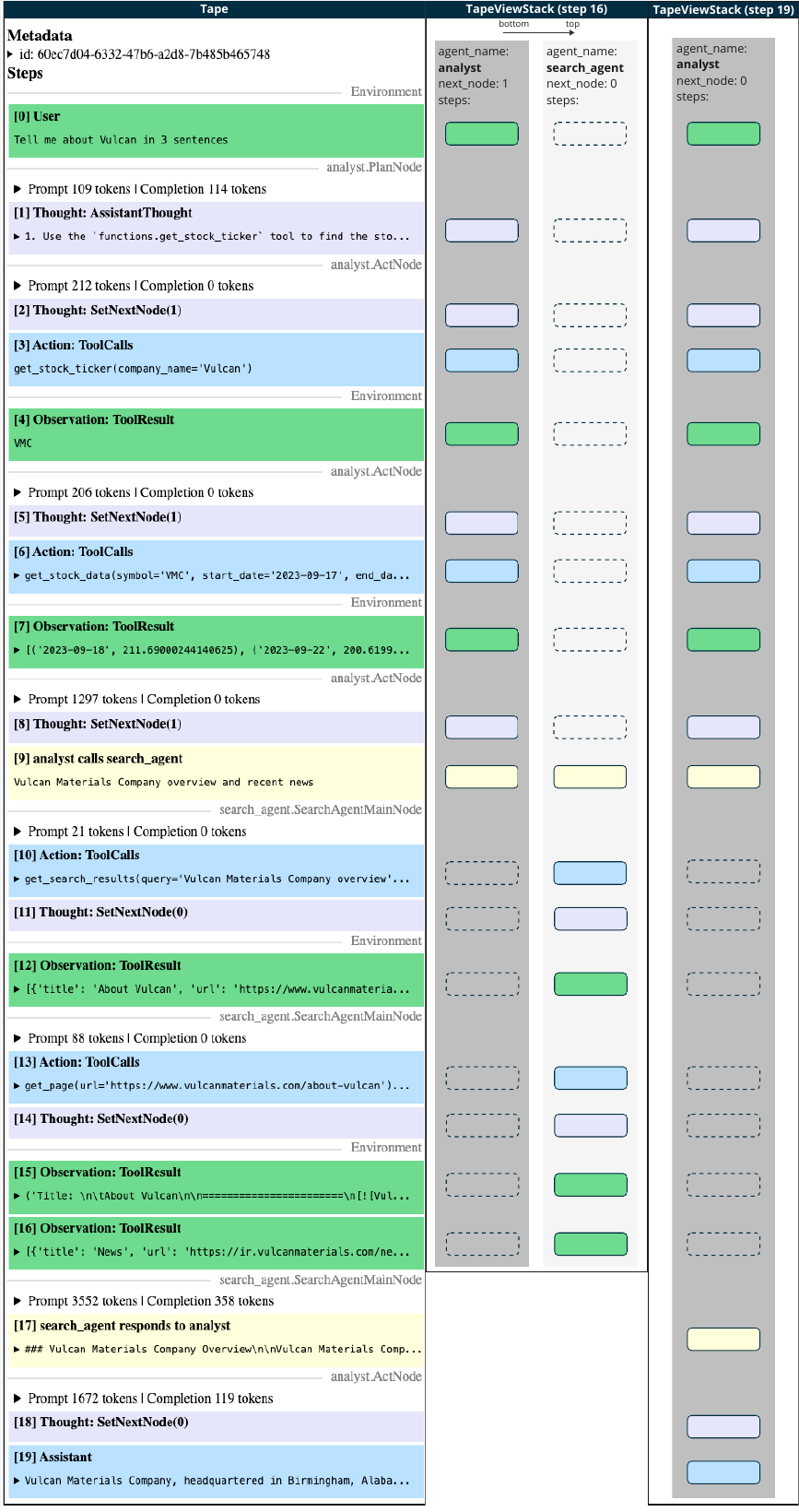}
  \end{minipage}
  \caption{A multi-agent tree structure (left) and a tape resulting from their work (middle) with the TapeViewStack at specific steps (right). At step 16 the stack's top view is the SearchAgent’s tape view. At step 19, only the Analyst's view exists. Note how the Analyst's view does not include the Search Agent's steps except for its response. Steps are color-coded: yellow for communication thoughts, purple for internal agent thoughts, blue for actions, and green for observations. The step’s author is indicated in grey using the ``agent.node'' format. Appendix~\ref{appendix:multi_agent_code} shows the code implementation that produced this tape.
  % SetNextNode is never part of views
  }
  \label{fig:view_agent_run}
\end{figure}

The metadata is crucial for building the tooling and the algorithms that empower the agent developer. \Cref{fig:view_agent_run} shows a visualization of some metadata fields.

\section{TapeAgents: tooling}
\label{sec:tapeagents_tooling}
The TapeAgents foundation that we covered in \cref{sec:tapeagents_foundations} allows the creation of a wide range of reusable agent components, tooling and learning algorithms. What the right building blocks and tooling are often depends on the application area. In our initial release, we provide several prototypes to jump-start future open-source collaborations.
% Finally, we cover the low-code paradigms, the development and the agent optimization tooling that one can build on the TapeAgent core foundations.

\subsection{Low-code Mini-Frameworks}
\label{subsec:mini_frameworks}
Building agents requires implementing many similar template rendering (\texttt{node.make\_prompt}) and text parsing (\texttt{node.generate\_steps}) routines. As a part of TapeAgents, we provide three examples of low-code mini-frameworks for building agents by composing and configuring off-the-shelf components:
\begin{enumerate}
\item \textbf{MonoAgent} exemplifies the most straightforward way to implement a monolithic agent: make a comprehensive prompt from all the data from the tape and the possible step schemas, then parse the LLM output using the schemas. One creates a MonoAgent from MonoNode nodes whose prompts are the same except for the final user message instruction. A MonoAgent also requires the agent developer to provide Pydantic models for all possible steps that the agent can generate.
\item \textbf{TeamAgent} shows how an AutoGen-style agent team can work in one tape. One can create three different kinds of team agents: (a)~an initiator that send the first \texttt{Call} message, (b)~a manager that chooses the next active agent, (c)~a worker agent that responds using its system prompt. 
\item \textbf{LLMFunction} demonstrates how one can build agents using function-style prompt templates, akin to DSPy signatures. These prompt templates are particularly easy to optimize by adding demonstrations.
\end{enumerate}
We include the mini-frameworks mostly for demonstration purposes, as it is hard to offer a high-level programming paradigm without a good knowledge of the intended application domain. The TapeAgents paradigm makes it easy to build such mini-frameworks thanks to the agent's double compositionality (agents and nodes).
\subsection{Tooling}

In TapeAgents, the agent configuration and the tape are highly structured and linked with metadata. This allows us to offer developer tooling for a broad range of possible TapeAgents. In the initial release, we include several app prototypes. We offer \textit{TapeAgents Studio} (see \cref{fig:studio} in \cref{appendix:tools}), an app to interact with the agent and its tape, \textit{Tape Browser} (\cref{fig:tape-browser} in \cref{appendix:tools}), an app to inspect a batch of tapes, and \textit{Tape Diff} (\cref{fig:tape-diff} in \cref{appendix:tools}), which compares two batches of tape.  Furthermore, for agent optimization, we provide algorithms for auto-prompting, LLM fine-tuning, and a modular Reinforcement Learning orchestrator. The finetuning component uses \texttt{Accelerate}~\citep{accelerate} and ~\texttt{DeepSpeed}~\citep{Rasley2020DeepSpeedSO} libraries and supports tuning resumption, experiment tracking, reproducibility, LoRA tuning, and distributed training. The above apps and algorithms represent the first steps towards the fully fledged Studio and Optimize modules that we envision in \cref{fig:overview}.

\section{Examples}
In an initial set of examples, we demonstrate agents that represent different agent-building paradigms, as well case-studies of using different agent optimization methods. 

\label{sec:examples}
\subsection{Financial Analyst and Their Web Search Helper}
To offer an example with the maximal educational value, we have implemented a user-facing financial analyst agent that can delegate searching the web to its subagent. We show the structure of the analyst agent and an example tape in Figure~\ref{fig:view_agent_run}. Our introductory hands-on notebook\footnote{\url{https://github.com/ServiceNow/TapeAgents/blob/main/intro.ipynb}} takes the reader through a journey from TapeAgents basic concepts to building this agent. 

For illustrative purposes, we implemented the nodes in this example from scratch, without using mini-frameworks from Section~\ref{subsec:mini_frameworks}. We offered the analyst agent an environment with several tools: one to get the company ticker, another to download stock data, as well as several tools to search and browse the web. We inform the analyst and their web search helper of the tools that they can use by including their tools' schemas in the prompts that the respective agent's nodes make. The agent operates on a tape type called \texttt{DialogTape}, which can only contain two kinds of actions: \texttt{ToolCalls} to call one or more tools and \texttt{AssistantStep} to respond to the user. 
The agent uses the same \texttt{ToolCalls} step with different content to call different tools. This is the use we intend for \texttt{DialogTape}: quick agent prototyping without declaring usecase-specific step schemas, though we believe most TapeAgents users will find it useful to declare their own action and thought types.

\subsection{Open-domain Question Answering and Web Browsing With Monolithic Agents}
\label{sec:gaia_workarena}
\begin{figure}
    \centering
    \includegraphics[width=0.7\linewidth]{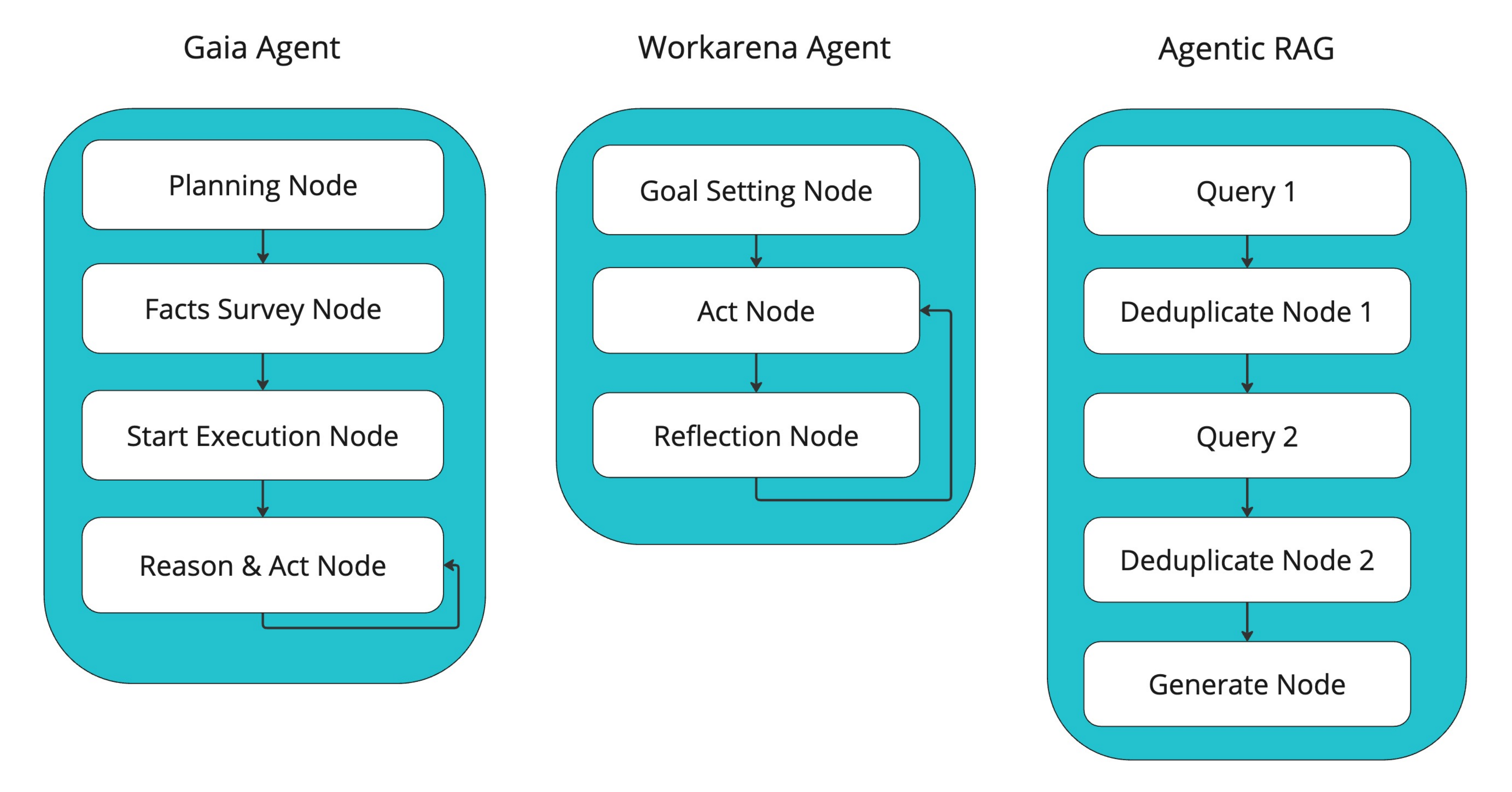}
    \caption{Agent structures for GAIA, WorkArena and Agentic RAG experiments (see sections 4.2 and 4.5 for details).}
    \label{fig:gaia_workarena_rag}
\end{figure}

To validate TapeAgents quantitatively, we build two agents that target existing benchmarks. The first one is a question-answering (QA) agent that targets the GAIA benchmark~\citep{mialon2024gaia}. The QA agent can search the web, run Python code, read multiple file types. To meet the GAIA evaluation requirements, we prompt the agent to output the precise short answer only. We build the QA agent from \texttt{MonoNode} nodes, with two planning nodes and one acting node in which the agent loops (see \cref{fig:gaia_workarena_rag}). \Cref{tab:gaia} shows that the agent performs quite well for such a simple agent. Our agent beats the more complicated Magentic-1 agent ~\citep{fourney2024magentic} which incorporates Orchestrator and multiple executor agents and also progress ledger. For comparison implementing our agent only requires gathering the tools, declaring corresponding action steps (like \texttt{ReadDocumentAction} and \texttt{UseCalculatorAction}) and declaring usecase-specific thoughts for reasoning (like \texttt{ListOfFactsThought} and \texttt{NewFactThought}). There is a lot of room for further improvement of the agent, starting from the obviously beneficial majority voting ensembling of runs and up to the multiagents, self-critic techniques, search over the tree of thoughts, etc.

We used a similar approach to build a web-browsing agent that targets the WorkArena benchmark~\citep{drouin2024workarena}. The BrowserGym environment has been used for the evaluation, with a convenient interface to the real browser. We declared action classes for different browser actions like \texttt{HoverAction} and \texttt{PressAction} and so on. \Cref{fig:gaia_workarena_rag} illustrates the exact agent structure, which is following ReAct~\citep{yao2023reactsynergizingreasoningacting}+Planning approach, with goal-setting, reflection and acting nodes.  We benchmark our web agent and find that it performs competitively (see \cref{tab:workarena}).


\begin{table}[]
\caption{GAIA Agent evaluation results.}
\label{tab:gaia}
\centering
\begin{tabular}{lcc}
\toprule
\textbf{Framework \& Model}                  & \textbf{Val accuracy, \%} & \textbf{Test accuracy, \%}          \\ \cline{1-3}
\cellcolor[HTML]{EFEFEF}Langfun Agent v2.0, (Gemini, Claude Sonnet 3.5)    & \cellcolor[HTML]{EFEFEF}54.5            & \cellcolor[HTML]{EFEFEF}49.3  \\
\cellcolor[HTML]{FFFFFF}HuggingFace Agents, GPT-4o    & \cellcolor[HTML]{FFFFFF}44.2            & \cellcolor[HTML]{FFFFFF}33.3  \\
\cellcolor[HTML]{EFEFEF}TapeAgent, GPT-4o                & \cellcolor[HTML]{EFEFEF}37.0          & \cellcolor[HTML]{EFEFEF}33.2       \\
\cellcolor[HTML]{FFFFFF}MSR Magentic-1, GPT-4o                & \cellcolor[HTML]{FFFFFF}36.9          & \cellcolor[HTML]{FFFFFF}32.3       \\
\cellcolor[HTML]{EFEFEF}TapeAgent, GPT-4o-mini           & \cellcolor[HTML]{EFEFEF}27.3          & \cellcolor[HTML]{EFEFEF}21.9      \\
\cellcolor[HTML]{FFFFFF}GPT-4 + manually selected plugins & \cellcolor[HTML]{FFFFFF}14.6          & \cellcolor[HTML]{FFFFFF}14.6    \\
\bottomrule
\end{tabular}
\end{table}

\begin{table}[]
\caption{Workarena Agent evaluation result on Workarena L1 tasks.}
\label{tab:workarena}
\centering
\begin{tabular}{lc}
\toprule
\textbf{Framework \& Model} & \textbf{Accuracy, \%}   \\ \cline{1-2}
\rowcolor[HTML]{EFEFEF} 
TapeAgents GPT-4o & 44.2  \\
\rowcolor[HTML]{FFFFFF} 
Agentlab GPT-4o & 42.7  \\
\rowcolor[HTML]{EFEFEF} 
TapeAgents GPT-4o-mini & 29.1  \\
\rowcolor[HTML]{FFFFFF} 
Agentlab GPT-4o-mini & 23.0 \\
\bottomrule
\end{tabular}
\end{table}

\subsection{Data Science With a Team of Agents}
To demonstrate that TapeAgents natively supports the multi-agent paradigm, we implement a ``data science'' agent team that consists of the Requestor, Manager, Software Engineer, Code  Executor and Asset Reviewer agent. Figure \ref{fig:multi-agent} (in Appendix \ref{appendix:tape}) shows the team in action as it builds a stock price comparison plot. We drew inspiration from the popular AutoGen framework for the multi-agent communication pattern in this example. Benefits of the TapeAgents implementation of this agent team include that one can easily resume the team from an intermediate tape or use tapes to optimize the entire agent organization algorithmically.

\subsection{Finetuning a Cheap Math Agent}
We test-drive TapeAgents fine-tuning component with a distillation example. We train a LLAMA3.1-8B-based math agent using tapes by its teacher counterpart with LLAMA3.1-70B under the hood. We equip each agent with a reasoning node and run the environment with a calculator tool. After finetuning on 3,000 samples from 1,000 teacher tapes, the student performance rises significantly from 66.2\% to 77.5\%, though the teacher's performance, at 93.1\%, remains much higher still.

\subsection{Prompt-Tuning for Agentic RAG}
In our last example, we show how the tape, the agent configuration and the metadata linking them, can serve as a medium to implement data-driven agent optimization algorithms. In this example, we use \texttt{LLMFunction}
% TODO: link another section 
prompt templates (see examples in Appendix~\ref{app:llm-function-templates}) that describe the intended behavior of a transformation that the LLM should perform, including the instruction, the input/output format, and optionally, a few demonstrations.
% TODO: add prompt example to appendix
We designed \texttt{LLMFunction} to make it possible to implement DSPy-like algorithms in TapeAgents. Below we describe how we used TapeAgents components to closely reimplement the DSPy introductory notebook. 

We compose a Retrieval-Augmented Generation (RAG) agent. \Cref{fig:gaia_workarena_rag} illustrates the structure where the agent performs two rounds of query generation and Wikipedia retrieval and then produces a short factual answer.
% TODO: cite Baleen paper
We build this agent mostly from \texttt{LLMFunctionNode} nodes that describe how the input fields in their respective \texttt{LLMFunction} templates should be filled with the steps from the tape. The only different kind of a node is a null-prompt node that deduplicates the retrieved paragraphs. We tune the prompts of the resulting 5-node agent by adding demonstrations to the function prompt templates.
% Dima: I prefer to have these references in other places in the text.
% See Appendix~\ref{app:llm-function-templates} for an example of those templates. The \texttt{add\_demos} function (Appendix~\ref{app:add-demos}) exemplifies that process. 
We obtain demonstrations by running the agent on 50 HotpotQA training examples and filtering the tapes with the wrong answer or duplicate queries. We try 10 different combinations of 4 randomly selected good tapes, extract demonstrations from them, add them to the agent's prompt template and measure the validation set performance, namely the sum of retrieval and answer accuracies. We select the best agent and evaluate on a set of 300 examples. Results in \cref{tab:agent-optimization} show that prompt-tuning leads to significant gains in both retrieval and answer accuracy.
%retaining the prompts that achieves the highest mean accuracy for both retrieval and answering on 100 validation examples. In this setting, prompt-tuning leads to significant gains in retrieval accuracy, which results in considerable gains in answer accuracy, as evaluated on 300 testing examples (see Table ~\ref{tab:agent-optimization}).
The optimized agent is still a TapeAgent that can be resumed from any intermediate tape, unlike a free-form Python program that uses DSPy. Notably, the implementation of the actual demonstration selection algorithm took just 12 lines of code (see Appendix~\ref{app:add-demos}), highlighting how the TapeAgent structures and metadata facilitate algorithm implementation. 

\begin{table}[]
\caption{Agent optimization results on HotpotQA.}
\label{tab:agent-optimization}
\centering
\begin{tabular}{lcccc}
\toprule
\textbf{Model} &\textbf{Agent} & \textbf{Retrieval (\%)} & \textbf{Answer (\%)} & \textbf{Mean Accuracy (\%)}  \\ \cline{1-5}
\rowcolor[HTML]{EFEFEF} 
GPT-3.5-Turbo & Non-optimized & 42.7 & 44.7 & 43.7  \\
\rowcolor[HTML]{FFFFFF}
              & Optimized & \textbf{56.7} & \textbf{47.0} & \textbf{51.8} \\
\rowcolor[HTML]{EFEFEF} 
\midrule
GPT-4o-mini & Non-optimized & 50.3 & 47.3 & 48.8 \\
\rowcolor[HTML]{FFFFFF}
            & Optimized & \textbf{57.0} & \textbf{50.7} & \textbf{53.8} \\
\rowcolor[HTML]{EFEFEF} 
\midrule
Llama-3.3-70B-Instruct & Non-optimized & 53.7 & 35.0 & 44.3 \\
\rowcolor[HTML]{FFFFFF} 
& Optimized & \textbf{57.7} & \textbf{48.3} & \textbf{53.0} \\
\bottomrule
\end{tabular}
\end{table}

\section{Case Study: Building a Cost-Effective Enterprise Form-Filling Assistant\label{sec:ghreatintro}}
A key use case of TapeAgents is optimizing LLM Agents to offer great quality services at a fraction of the cost. In this section, we present a fleshed-out example of how these goals can be achieved for a conversational assistant that can help fill a request form and submit the request.

% DIMA: I shortened this text, see new version above
% To exemplify the intended use of the TapeAgents framework, we present a case study on training a conversational form-filling bot. Many processes and administrative workflows in the enterprise world can be formulated as submitting a request to the relevant decision-makers.
%
% A first step towards improving access to these services is to deploy an AI agent which can help a user correctly identify and fill out the relevant form: reserve a flight, request time off, apply for benefits, report a lost badge, request a reimbursement, etc.
%
% To provide the best assistance, the AI agent should have access not just to the documentation available for the form, but also to search the relevant textual knowledge (i.e. company policies) and query the relevant systems of records (i.e. employee directory, database of reservations).

\subsection{Problem Setting} 
Employees in large enterprises often fill forms to request resources, assistance or access. A conversational assistant can make the form-filling experience smoother by guiding its user to the right form, by accepting the user's free-form inputs, and by answering the questions that the user may have in the process. For a great experience, the assistant must also gracefully handle the ``unhappy-path'' situations, such as when the user's ask is impossible to fulfill or when the assistant cannot answer the user's question. In this case study, we show one can use TapeAgents to train a cost-effective assistant that scores high according to a formal metric of user experience that we call the {\em GREADTH} score. GREADTH stands for Grounded, REsponsive, Accurate, Disciplined, Transparent, and Helpful. We will explain these metrics in Section~\ref{subsec:ghreat_metrics}.

For simplicity, we consider building a restricted assistant:
\begin{itemize}
    \item The assistant should answer questions solely based on the form documentation; it does not have to retrieve any additional documents.
    \item The assistant can only help with one form at a time.
    \item At the start of the conversation the assistant converses with the user to guide them to the correct form.
    \item After the form is chosen, the assistant will help the user fill out the form correctly. The agent will not allow the user to switch to a different form after this point. 
    \item During the form-filling process, the assistant maintains the field values that the user has provided so far.
    \item The interaction ends with either the form submission or the agent exiting the conversation after the user's confirmation.
\end{itemize}
The resulting form-filling setup is reminiscent of the Task-Oriented Dialogue setting that has been widely discussed in the literature~\citep {rastogi2020towards,budzianowski2018large}. Following this body of work, we will refer to form fields as \textit{slots}.

\subsection{Evaluation Criteria: GREADTH Experience}
\label{subsec:ghreat_metrics}

Despite the apparent simplicity of the form-filling setup, it can be non-trivial to develop a form-filler assistant that balances an excellent conversational experience with low hallucination rate and reasonable cost. To balance these desiderata, one must first define them in a measurable way. In our case study, we design our assistant to have maximum \textbf{GREADTH}: \textbf{G}rounded, \textbf{RE}sponsive, \textbf{A}ccurate, \textbf{D}isciplined, \textbf{T}ransparent, and \textbf{H}elpful. We define these aspects as follows:
\begin{itemize}
    \item Everything a \textbf{G}rounded assistant says must be fully supported by the form documentation, the conversation history and the \textit{grounding statement}. The latter defines the assistant's identity and purpose and constrains the assistant to form-filling. Small talk is considered ungrounded.
    \item A \textbf{H}elpful assistant must actively take the conversation forward by asking for the user's intent, requesting the next slot to fill, or asking for confirmation before making the request once all slots have been filled. It should also (a)~provide all relevant information regarding a slot when asking for it (default value, allowed values, optionality), (b)~answer any user question if the form documentation provides the relevant information, (c)~exit the conversation at any time if the user desires so.
    \item An \textbf{A}ccurate assistant must correctly identify the user's intent and import the relevant form documentation, fill the slots correctly based on user messages, update the slots if the user changes their mind, or skip the slots if relevant.
    \item A \textbf{T}ransparent assistant acknowledges all changes made to the partially filled form. This includes \textit{slot-filling} or \textit{skipping} slots that are optional or have a default value. The summary of slots changes can be concise, yet the user must be able to understand how the slots were affected. While in a mixed modality interaction the user may visually see the form changes, in a purely voice interaction such as talking over the phone the transparent behavior is essential.
    \item A \textbf{D}isciplined assistant must follow its planning thoughts, such as requesting a slot, asking for confirmation, answering a question, rejecting incorrect slot values (as defined by the form documentation), or rejecting an invalid ask.
    \item We require the assistant to be \textbf{RE}sponsive to address a common experience issue with AI assistants: the robotic and opaque behavior when the user goes off the expected conversational path. We wish that AI assistant infers and acknowledges what the user had in mind, while explaining that their request or question is not possible. In particular, we want to cover the following scenarios:
    \begin{itemize}
        \item if the user tries to fill a slot with an invalid value, the assistant should acknowledge the value and respond that it is invalid;
        \item if the user offers information that looks like a value for a nonexistent but plausible slot, the assistant should acknowledge the value and the inferred name of the slot and respond that such slots is not available in this form;
        \item if at the form-filling stage the user's ask looks like a request for another form, the agent should acknowledge their ask and say that it can not fulfill it right now (note that in our setup the user must either finish filling the current form or exit);
        \item if the user asks a plausible question that the form documentation does not answer, the agent should acknowledge the question and say that it can not be answered.
    \end{itemize}
    The user may ask other requests that have nothing to do with form-filling (i.e. weather requests). In that case, the agent must politely decline the request and keep moving the conversation towards either submitting or aborting the request.
    To align our definition of responsiveness with the common sense meaning of this word, we also require a \textbf{RE}sponsive assistant to acknowledge all valid slot values and valid questions. Thus we often deem a response that is not \textbf{T}ransparent, or not \textbf{H}elpful also not \textbf{RE}sponsive.
\end{itemize}
The GREADTH criteria above are binary, a conscious choice that we made to simplify the evaluation and the analysis. We acknowledge that this makes them somewhat crude, as e.g. two assistant answers can be both technically correct but can widely differ in readability and in the choice of the information to present. One can complement these criteria with a preference-based experience evaluation that implicitly covers fluency, verbosity, and other aspects of the assistant's response.

\subsection{Design of a Form Filling TapeAgent}
\label{subsec:ghreat_design}

% \paragraph{GREADTH Explainability}

\paragraph{Tape Structure}

% A common issue with language models is the lack of explanability of the model predictions. Approaches such as chain-of-thought~\citep{wei2022chain} and ReAct~\citep{DBLP:conf/iclr/YaoZYDSN023} have been proposed to mitigate this. In the case of the form-filler agent, we take it one step further by structuring the agent tape as a succession of internal thoughts, actions, and observations.
%%%% To address Dima's feedback on focusing on planning

To build an agent that provides a GREADTH experience, we decompose the conversational form-filling task into smaller reasoning steps before each agent message. Having the GREADTH metrics in mind, we define the agent's thoughts to help it plan its response.
The thoughts are used to represent a chain-of-thought of the agent, which includes (a)~analyzing the user's intent (e.g. the form that is requested, the provided slot values, a question being asked); (b)~updating the internal state of the conversation (e.g. slot-filling); and (c)~planning the next actions (e.g. requesting a specific slot value or requesting a user confirmation).
In particular, after each user message or observation, the agent must return:
% such as \texttt{RefuseInexistentFunctionThought}, \texttt{AnswerFromFunctionSchemaThought}, \texttt{RequestFunctionParametersThought}, etc.
% Form identification is an action
% Slot-filling is also defined as a thought (\texttt{UpdateFunctionParametersThought}) because it does not require intervention from the Environment. The full list of planning thoughts available to our Agent is described in Appendix~\ref{appendix:ghreatthoughts}.
% 
% Following common practice\citep{wei2022chain,DBLP:conf/iclr/YaoZYDSN023}, we decompose the form-filling task into reasoning traces and plans. The chain-of-thought paradigm equips the model with observability by demonstrating the model how to reach a particular outcome and facilitates debugging to identify where exactly errors occur. In the case of the form-filler agent, we take this paradigm one step further by structuring the agent tape as a succession of internal thoughts, actions, and observations.
% 
%
% Taking inspiration from the \textit{System Acts} defined in the Schema-Guided Dialogue Dataset or SGD~\citep{rastogi2020towards}, we define a list of possible \textit{internal thoughts} to help structure the agent's chain-of-thought. After each observation, the agent must return:
\begin{enumerate}
    \item A list of \textbf{thoughts}: these include slot-filling related thoughts such as \texttt{UpdateFunctionParametersThought}\footnote{While working on the form-filling case-study we interchangeably used the terms form / request / function and the terms slot / field / parameter. In this report we use the original technical names for the tape's steps to maximize the coherence between the text and the code.}; and message planning thoughts specifying the next slot to request (\texttt{RequestFunctionParametersThought}), the need to ask for confirmation before submitting the request (\texttt{RequestFunctionCallConfirmationThought}), planning to answer a question (\texttt{AnswerFromFunctionSchemaThought}), planning to inform the user that their question cannot be answered (\texttt{NoAnswerInFunctionSchemaThought}), refusing unsupported request/behavior/slot values (\texttt{RefuseInexistentFunctionThought} / \texttt{RefuseToEngageThought} / \texttt{RefuseInvalidFunctionParameterValueThought}), etc. The full list of thoughts is described in Appendix~\ref{appendix:ghreatthoughts}.
    \item A single \textbf{action}: the agent returns a single action, such as searching available forms (\texttt{ResolveFunctionAction}), retrieving form documentation (\texttt{InspectFunctionAction}), replying to the user (\texttt{PromptUserForTextMessageAction}), submitting the request (\texttt{CallFunctionAction}), or exiting the conversation (\texttt{ExitAction}). Each action results in new observations (available forms, retrieved documentation, or user input), and ends the current agent turn.
\end{enumerate}

\begin{figure}[t]
    \centering
    \includegraphics[width=\linewidth]{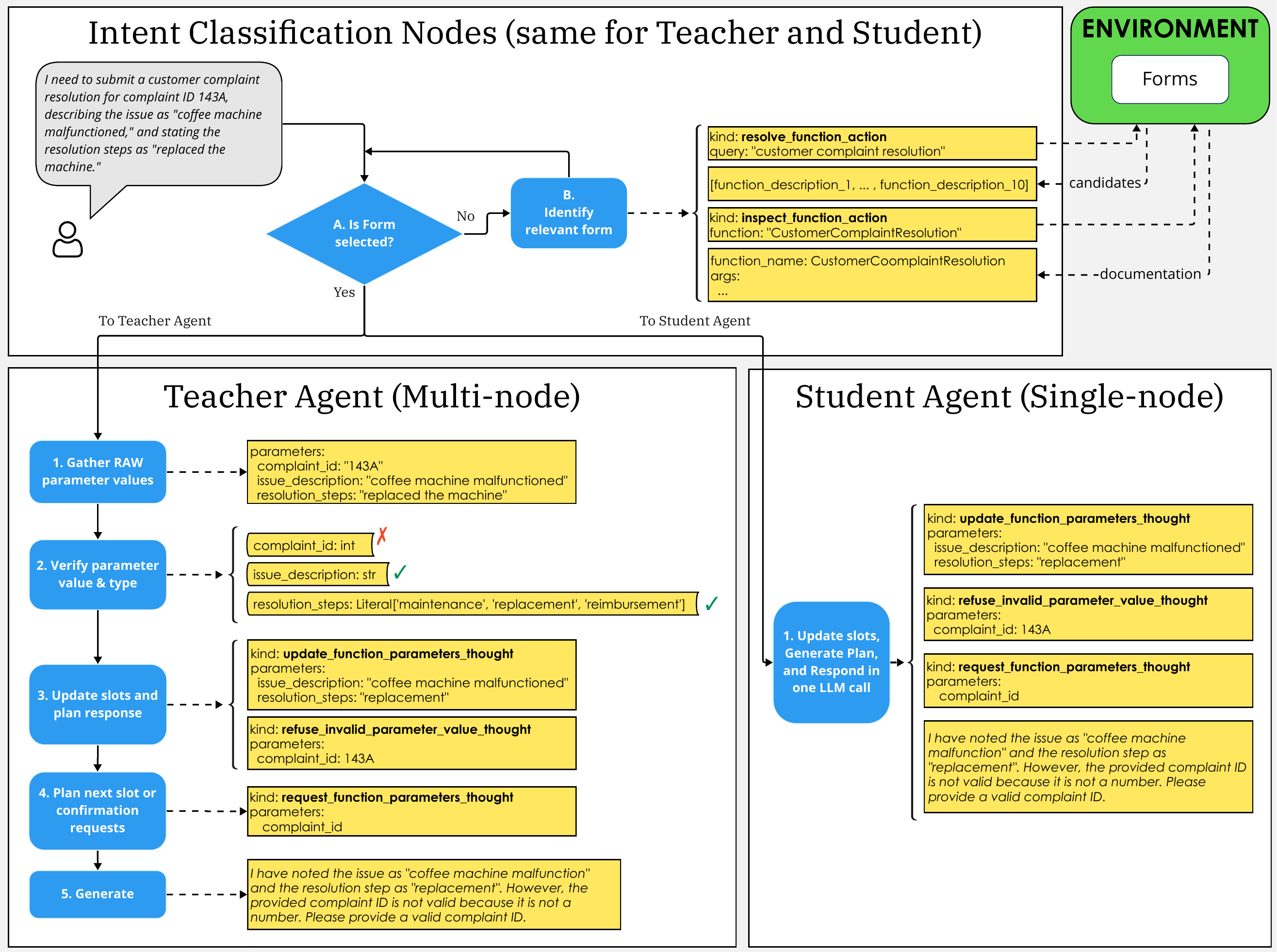}
    \caption{Node structure of the Teacher and the Student agents. The Teacher agent combines intent classification nodes and 5 extra nodes, while the Student agent combines intent classification nodes plus 1 additional node. We represent nodes in blue and produced steps in yellow. Please note that words \textit{intent/form/function} and \textit{slot/field/parameter} are used interchangeably in this report.}
    %\caption{Teacher, student, and intent classification prompts. The teacher agent combines intent classification and teacher prompts, while the student agent combines intent classificaiton and student prompts. We represent nodes in blue and produced steps in yellow. Please note that words \textit{intent/form/function} and \textit{slot/field/parameter} are used interchangeably between code and paper.}
    \label{fig:multistep_teacher}
\end{figure}

\paragraph{Multi-node Teacher Agent}
We experimented with various combinations of prompting techniques and LLMs to obtain the best performance (see GREADTH metrics in Sections~\ref{subsec:ghreat_metrics} and human evaluation procedure in Section~\ref{subsec:ghreat_experiments}). We found that using Llama-405B with multi-step prompting\footnote{Also known as prompt chaining or step-wise prompting. More information can be found in prompt engineering guidelines, e.g. \url{https://docs.anthropic.com/en/docs/build-with-claude/prompt-engineering/chain-prompts} and \url{https://platform.openai.com/docs/guides/prompt-engineering/tactic-specify-the-steps-required-to-complete-a-task} (Accessed on Oct 15, 2024)} (multiple TapeAgent nodes) yielded the most promising results based on human evaluation. For additional details about the choice of model and agent structure, see Appendix~\ref{appendix:teachers}.
Our Teacher agent is made of 7 nodes: 2 for intent classification and 5 for slot filling (Figure~\ref{fig:multistep_teacher}):
\begin{itemize}
    \item \textbf{Is Form Selected?} The initial node checks if we identified the user's intent. This node does not call any LLM, it simply checks if an \texttt{InspectFunctionAction} step is already present in the current Tape. If a form has not yet been selected, the model proceeds to the next step (Step (B) in Figure~\ref{fig:multistep_teacher}). Otherwise, the model moves directly to the form-filling phase (Step 1 (left) in Figure~\ref{fig:multistep_teacher}).
    \item \textbf{Intent Discovery}. This node (a)~lists the available forms (\texttt{ResolveFunctionAction} queries the Environment to return a list of available forms); and (b)~finds the relevant one based on the previous user's message (\texttt{InspectFunctionAction} queries the Environment to return the form documentation). If the LLM cannot identify the relevant form, it is prompted to ask again the user for its intent (\texttt{RefuseInexistentFunctionThought; RequestFunctionThought; PromptUserForTextMessageAction}).
    \item \textbf{Gather Raw Parameter Values}. Once the user's intent is discovered and its form documentation imported, the LLM is prompted to extract all raw slot values present in the user's message and yield a \texttt{GatherValuesThought} step.
    \item \textbf{Verify Parameter Values \& Types}. In this node, the LLM is prompted to verify all extracted slot values based on the form documentation imported. The LLM verifies that each value is of the correct type and is valid (in the case of categorical slots) with a \texttt{VerifyValuesThought} step.
    \item \textbf{Update Slots and Plan Response}. Once we have identified the correct/incorrect slot values, the LLM is prompted to update the filled slots with a \texttt{UpdateFunctionParametersThought} step, refuse invalid slots with \texttt{RefuseInvalidFunctionParameterValueThought} steps, and start planning its response to the user with \texttt{AnswerFromFunctionSchemaThought} if applicable.
    \item \textbf{Plan Next Slot or Confirmation Requests}. This node prompts the LLM to move the conversation forward by either (i)~requesting the next slot (\texttt{RequestFunctionParametersThought}), (ii)~confirming that the user wants to submit the current request (\texttt{RequestFunctionCallConfirmationThought}), or~(iii) confirming that the user wants to exit the chat (\texttt{RequestExitConfirmationThought}).
    \item \textbf{Generate}. Eventually, the final node of our Teacher TapeAgent prompts the LLM to generate the next action. Based on the previous steps/thoughts in the Tape, the agent can either (i)~write a message (\texttt{PromptUserForTextMessageAction}), (ii)~submit the request (\texttt{CallFunctionAction}), or (iii)~exit the conversation (\texttt{ExitAction}).
\end{itemize}
See Figure~\ref{fig:teacher_tape} of Appendix~\ref{appendix:ghreat_tapes} for a sample Teacher tape.

\paragraph{Single-node Student Agent} We design the Student agent with the goal of minimizing input and output token counts in order to optimize cost. We describe the slot-filling task in one condensed LLM prompt (1 TapeAgent node) and use Llama-8B as the LLM to the Student agent to further optimize cost and inference time.
While we could get rid of the instructions entirely in the LLM prompt, we still provide enough instructions so that a strong model (e.g. Llama-405B) can still make sense of the task in a zero-shot setting.
Our Student agent is made of 3 nodes: 2 for intent classification and 1 for slot filling (Figure~\ref{fig:multistep_teacher}):
\begin{itemize}
  \item \textbf{Is Form Selected?} Same structure as the Teacher but with fewer and more compact instructions.
  \item \textbf{Intent Discovery}. Same structure as the Teacher but with fewer and more compact instructions.
  \item \textbf{Update Slots, Generate Plan, and Respond}. Using a single short prompt, we ask the agent to generate all its thoughts (e.g. \texttt{RefuseInvalidFunctionParameterValueThought; UpdateFunctionParametersThought; RequestFunctionParametersThought}) and end with an action (e.g. \texttt{PromptUserForTextMessageAction}).
\end{itemize}
We show an example of the Student tape in Figure~\ref{fig:student_tape} of Appendix~\ref{appendix:ghreat_tapes}.

%
% A widely-used framework to study conversational assistants is Helpful-Honest-Harmless decomposition by~\citet{bai2022training}. In Figure~\ref{figure:relationmetric}, we illustrate how our criteria relate to those from the HHH framework.

% \begin{figure}
% \label{HHH-vs-GREADTH}
% \centering
% \includegraphics[width=0.8\linewidth]{figures/hhh_to_ghreat.pdf}
% \caption{Relation between HHH~\citep{bai2022training} and GREADTH criteria. A Grounded agent will be HHH-Honest and most often HHH-Harmless, as the form documentation will not typically contain harmful information that the assistant should not tell the user. Our other GREADTH metrics can be seen as setup-specific aspects of the broad HHH-Helpfulness. Responsiveness overlaps with Transparency and Helpfulness. \label{figure:relationmetric}}
% \end{figure}

\subsection{Experiments\label{subsec:ghreat_experiments}}

We consider the task of training a cost-effective conversational agent to help the user fill and submit a form. We seek the conversational experience to score high in GREADTH metrics (Section~\ref{subsec:ghreat_metrics}), while requiring fewer tokens per interaction and less cost per processed token.
We attempt to distill a multi-node Teacher agent that uses a large model into a single-node Student agent that uses a small model. For both finetuning and evaluation purposes, we simulate the environment (synthetic forms and user interactions).

\paragraph{Synthetic Companies} 
To simulate an enterprise environment with multiple forms, we prompt a Llama-3-70B-Instruct model to generate the name and descriptions of 6 fictitious companies, which we divide into training domains (\texttt{FlyCorp, BigBankCorp, CoffeeCorp}) and testing domains (\texttt{DriveCorp, LuxuryCorp, ShopCorp}). For each company, we then prompt Llama-3-70B-Instruct to generate 10 plausible request forms based on the company name and description. Each form has a name and a description. Eventually, for each generated form description, we prompt Llama-3-70B-Instruct to generate a FunctionSchema for that request form. FunctionSchemas are structured data representations including a name, a description, a JsonSchema describing slots to fill, and a JsonSchema describing the object returned once the form is submitted. Slots to fill have a name, description, type (categorical, date, email, string), optionality, possible values, and default values. The prompts used to generate synthetic companies are described in Appendix~\ref{appendix:synthetic_companies}. Each of our simulated environments has 10 available forms and identifying which form the user requests is part of the form-filling task we tackle.

\paragraph{User Agents} In the multi-turn dialogue setting, distillation is more complex than running the Teacher on a set of static contexts because of alternating agent and user turns. The user turns need to either be produced by a human or generated.
We define 19 different (single-node) User agents to generate the next user message by simulating a variety of user behaviors. Some user behaviors are ``easy'' such as ``\textit{answer the agent's question}'', while others can be quite adversarial such as ``\textit{provide a good value for slot X and a bad value for slot Y}'' or ``\textit{ask for something unrelated}''. All user agents and their respective behaviors are described in Appendix~\ref{appendix:user_behaviors}. Each of these User agents serves as a special instance of the Tape Environment that responds to the assistant after each \texttt{PromptUserForTextMessageAction} step.

\paragraph{Synthetic Dialogue Generation}
We generate a dataset of dialogues by repeatedly and alternatively running the Teacher agent and a User agent to generate the next turn from a set of partial conversations. The minimal partial conversation contains only the ``\textit{Hi, how can I help you?}'' Assistant message. User agents are sampled randomly after each \texttt{PromptUserForTextMessageAction} steps from the Assistant, and yield a user message Observation. The Assistant then continues the conversation until its next message.
The conversations end when they reach 18 turns (9 agent messages and 9 user messages), whenever the request is submitted or aborted, or if the agent fails to produce a valid continuation (e.g. trying to fill a nonexistent slot). We show an example of a full conversation between the Teacher agent and the User agent in Figure~\ref{fig:teacher_tape} (Appendix~\ref{appendix:ghreat_tapes}).
We structure the dataset as a tree of dialogues, while controlling for the width of the tree (beam search), the diversity of user behaviors, and the diversity of filled forms (requests). We use a beam size of width 500, up to 9 user turns and 9 agent turns. This results in roughly \textbf{13k train} and \textbf{13k test} agent continuations per synthetic company.

% Define color gradients for low (red) to high (green) values
\newcommand{\verylow}[1]{\cellcolor{red!50}#1}
\newcommand{\low}[1]{\cellcolor{red!30}#1}
\newcommand{\mediumlow}[1]{\cellcolor{yellow!30}#1}
\newcommand{\medium}[1]{\cellcolor{green!20}#1}
\newcommand{\high}[1]{\cellcolor{green!40}#1}
\newcommand{\veryhigh}[1]{\cellcolor{green!60}#1}

\begin{table}[t]
\caption{\textbf{GREADTH Form Filler experiment results}. The Teacher\textsuperscript{1} is a multi-node agent with Llama 3.1 405B Instruct FP8 as its LLM. The Student\textsuperscript{2} is a single-node agent with Llama 3.1 8b Instruct as its LLM. We also evaluate the multi-node agent with GPT-4o and with Llama 3.1 8B Instruct as its LLM, as well as the single-node agent with Llama 3.1 405B Instruct for comparison. The metrics are computed over 1524 partial dialogues from the test domains. Read full analysis in Section~\ref{subsec:ghreat_experiments}.}
\label{tab:factscore}
\centering
\begin{adjustbox}{width=1\textwidth,center}
% \begin{tabular}{p{5cm}ccccccc}
% \toprule
% \textbf{Agent}(Model+Prompt) & \textbf{G} & \textbf{H} & \textbf{Re} & \textbf{A} & \textbf{D} & \textbf{T} & 
% \begin{tabular}{c} \textbf{GREADTH Score} \\ (Human Raters)\end{tabular} \\
% \midrule
% \multicolumn{8}{c}{\textbf{\textit{Reference Comparison (GPT-4o-2024-08-06)}}} \\
% Teacher Prompt (0-shot) & \veryhigh{91.3\%} & \high{87.2\%} & \high{87.1\%} & \veryhigh{91.4\%} & \veryhigh{92.7\%} & \veryhigh{94.3\%} & \medium{\textbf{74.9\%}} \\
% \midrule
% \multicolumn{8}{c}{\textbf{\textit{Teacher Model (Llama-3.1-405B-Instruct)}}} \\
% \textbf{Teacher Prompt}\textsuperscript{1} (0-shot) & \high{89.8\%} & \high{86.5\%} & \high{85.0\%} & \high{87.9\%} & \veryhigh{91.6\%} & \veryhigh{92.5\%} & \medium{\textbf{75.8\%}} \\
% Student Prompt (0-shot) & \medium{74.2\%} & \mediumlow{61.9\%} & \medium{72.0\%} & \medium{76.8\%} & \mediumlow{67.3\%} & \medium{78.9\%} & \mediumlow{\textbf{43.2\%}} \\
% \midrule
% \multicolumn{8}{c}{\textbf{\textit{Student Model (Llama-3.1-8B-Instruct)}}} \\
% Teacher Prompt (0-shot) & \medium{75.5\%} & \mediumlow{60.3\%} & \mediumlow{57.7\%} & \medium{72.4\%} & \medium{74.0\%} & \medium{76.3\%} & \mediumlow{\textbf{36.6\%}} \\
% Student Prompt (0-shot) & \low{18.8\%} & \low{12.7\%} & \verylow{6.2\%} & \low{10.9\%} & \low{11.6\%} & \verylow{9.4\%} & \verylow{\textbf{2.0\%}} \\
% \textbf{Student Prompt}\textsuperscript{2} (finetuned) & \veryhigh{92.1\%} & \high{87.1\%} & \high{86.4\%} & \veryhigh{90.2\%} & \veryhigh{94.4\%} & \veryhigh{95.1\%} & \medium{\textbf{76.6\%}} \\
% \bottomrule
% \end{tabular}
\begin{tabular}{p{5.5cm}ccccccc}
\toprule
\textbf{Agent (LLM+Nodes)} & \textbf{G} & \textbf{Re} & \textbf{A} & \textbf{D} & \textbf{T} & \textbf{H} & 
\begin{tabular}{c} \textbf{GREADTH Score} \\ (Human Raters)\end{tabular} \\
\midrule
\multicolumn{8}{c}{\textbf{\textit{Reference Comparison (GPT-4o-2024-08-06)}}} \\
Multi-node (0-shot) & \veryhigh{91.3\%} & \high{87.1\%} & \veryhigh{91.4\%} & \veryhigh{92.7\%} & \veryhigh{94.3\%} & \high{87.2\%} & \medium{\textbf{74.9\%}} \\
\midrule
\multicolumn{8}{c}{\textbf{\textit{Llama-3.1-405B-Instruct}}} \\
\textbf{Teacher}\textsuperscript{1}: Multi-node (0-shot) & \high{89.8\%} & \high{85.0\%} & \high{87.9\%} & \veryhigh{91.6\%} & \veryhigh{92.5\%} & \high{86.5\%} & \medium{\textbf{75.8\%}} \\
Single-node (0-shot) & \medium{74.2\%} & \medium{72.0\%} & \medium{76.8\%} & \mediumlow{67.3\%} & \medium{78.9\%} & \mediumlow{61.9\%} & \mediumlow{\textbf{43.2\%}} \\
\midrule
\multicolumn{8}{c}{\textbf{\textit{Llama-3.1-8B-Instruct}}} \\
Multi-node (0-shot) & \medium{75.5\%} & \mediumlow{57.7\%} & \medium{72.4\%} & \medium{74.0\%} & \medium{76.3\%} & \mediumlow{60.3\%} & \mediumlow{\textbf{36.6\%}} \\
\textbf{Student}\textsuperscript{2}: Single-node (0-shot) & \low{18.8\%} & \verylow{6.2\%} & \low{10.9\%} & \low{11.6\%} & \verylow{9.4\%} & \low{12.7\%} & \verylow{\textbf{2.0\%}} \\
\textbf{Student}\textsuperscript{2}: Single-node (finetuned) & \veryhigh{92.1\%} & \high{86.4\%} & \veryhigh{90.2\%} & \veryhigh{94.4\%} & \veryhigh{95.1\%} & \high{87.1\%} & \medium{\textbf{76.6\%}} \\
\bottomrule
\end{tabular}
\end{adjustbox}
\end{table}

\begin{table}
\caption{\textbf{Agent Cost vs. Performance Tradeoff.} On the evaluation set, we report the average number of input and output tokens per agent turn, and the average cost per million agent turns, by multiplying the number of tokens with the retail price of the cheapest provider on Openrouter as of October 3rd 2024 (\$0.055 per million input/output tokens for Llama-3.1-8B-instruct and \$1.79 for Llama-3.1-405B-instruct). The cost per million agent turns is only \$85 for the Student vs. \$28,157 for the Teacher, which represents a \textbf{factor of 300} in savings. The cost for the Teacher agent could be mitigated by prefix caching, but most of the multi-node prompts are context-dependent. Read full analysis in Section~\ref{subsec:ghreat_experiments}.}
\label{table:cost-tradeoff}
\centering
\begin{adjustbox}{width=0.85\textwidth,center}
\begin{tabular}{lcccr}
\toprule
\textbf{Agent (LLM+Nodes)} & \makecell{\textbf{Input Tokens} \\ \textbf{/turn}} & \makecell{\textbf{Output Tokens} \\ \textbf{/turn}} & \makecell{\textbf{Cost} \\ \textbf{/1M turns}} & \makecell{\textbf{GREADTH} \\ \textbf{score}} \\
\midrule
\multicolumn{5}{c}{\textbf{\textit{Reference Comparison (GPT-4o-2024-08-06)}}} \\
Multi-node (0-shot) & 15,431 & 526 & \verylow{\$43,837} & \high{74.9\%} \\
\midrule
\multicolumn{5}{c}{\textbf{\textit{Llama-3.1-405B-Instruct}}} \\
\textbf{Teacher}: Multi-node (0-shot) & 15,189 & 541 & \low{\$28,157} & \high{75.8\%} \\
Single-node (0-shot) & 1,435 & 97 & \mediumlow{\$2,742} & \mediumlow{43.2\%} \\
\midrule
\multicolumn{5}{c}{\textbf{\textit{Llama-3.1-8B-Instruct}}} \\
Multi-node (0-shot) & 15,189 & 541 & \mediumlow{\$865} & \mediumlow{36.6\%} \\
\textbf{Student}: Single-node (0-shot) & 1,437 & 208 & \high{\$90} & \verylow{2.0\%} \\
\textbf{Student}: Single-node (finetuned) & 1,441 & 110 & \high{\$85} & \high{76.6\%} \\
\bottomrule
\end{tabular}
\end{adjustbox}
\end{table}

\begin{figure}
\centering
\includegraphics[width=0.9\textwidth]{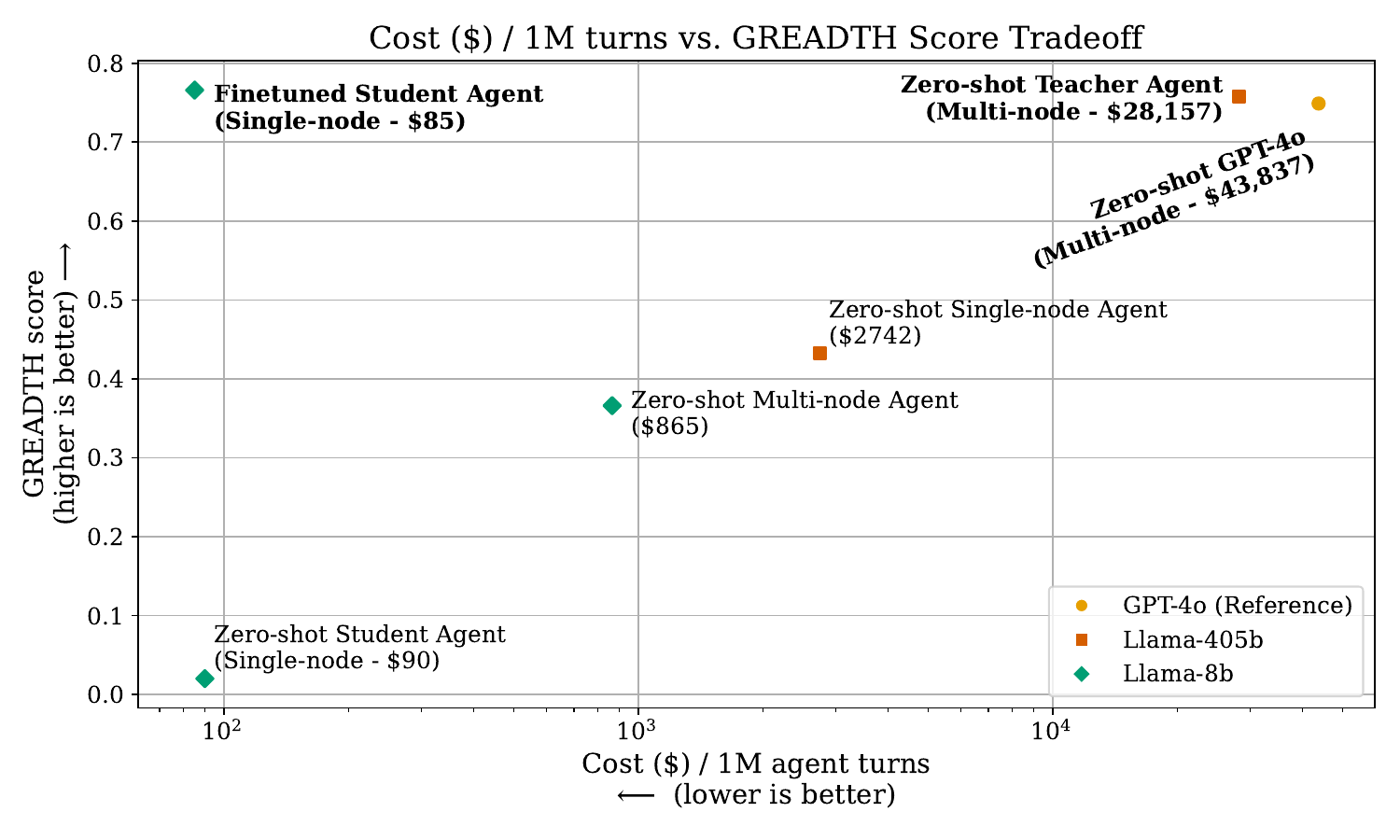}
\caption{\textbf{Cost per 1M Agent Turns vs. GREADTH Score Tradeoff}. The finetuned Student agent (top-left) performs on par with the Teacher and reference agents (top-right) for a fraction of the cost. We also provide ablations for various combinations of LLMs and node flows (single -vs- multi), which shows that finetuning is instrumental in getting the desired performance when using a short and simple LLM prompt (single-node). See Section~\ref{subsec:ghreat_experiments} for the full discussion.}
\end{figure}

\paragraph{Human Evaluation} We perform human evaluation to score each agent turn on the basis of the GREADTH metrics: groundedness, responsiveness, accuracy, discipline, transparency, and helpfulness (Section~\ref{subsec:ghreat_metrics}). The~labeling services were provided by Toloka,\footnote{\url{https://toloka.ai}} and the labelers were paid above the minimum wage in their respective work locations. Labelers provide the \textbf{6 GREADTH binary labels} per agent turn. We also have expert labelers who audit 10\% of the labels. We report scores for each metric as well as the \textbf{GREADTH score}, which is \textit{equal to 1 only if all 6 metrics are satisfied, and 0 otherwise} (binary AND). If the agent fails to produce a valid continuation --- due to unparsable JSON, invalid schema, attempting to fill an invalid slot value, or submitting a form with missing values --- then the agent automatically gets 0 for all 6 metrics.

\paragraph{Agent Distillation} As described in Section~\ref{subsec:ghreat_design}, we construct the \textbf{Teacher agent} by prompting a very large model, Llama-3.1-405B-Instruct-FP8 (runs on 8 H100-80GB GPUs) with sequential and lengthy instructions (\textbf{multi-node agent}) in order to obtain the best GREADTH score, with no regards to prompt length optimization or latency.
From a qualitative perspective, our interaction with the Teacher agent reveals a generally satisfactory user experience and validates its choice as a Teacher agent.
We construct the \textbf{Student agent} by combining a much smaller model, Llama-3.1-8B-Instruct model (runs on a single A100-80GB GPU), with a single-node (\textbf{single-node agent}) optimized for length.
We distill the Teacher agent into the Student agent by finetuning it over \textbf{13k teacher agent turns} (continuations) on the training domains. We perform a single epoch of LoRA~\citep{hu2021LoRA} optimization using AdamW~\citep{loshchilov2017decoupled} with learning rate $10^{-5}$ and batch size 32. Additional epochs did not seem to help.

\paragraph{Results} We evaluate the Teacher and Student agents by generating a single agent turn (continuation) over a set of \textbf{1524 partial dialogues} subsampled from the testing domains generated previously. We control for the diversity of user behaviors (19 mostly uniform) and for the diversity of requested forms (30 forms split over 3 domains + dialogues where no form has been selected yet). Here is what we observed:
\begin{enumerate}
    \item The Teacher agent achieves a GREADTH Score of \textbf{75.8\%} (Table~\ref{tab:factscore}) at the expensive cost of \textbf{\$28,157/1M} agent turns (Table~\ref{table:cost-tradeoff}), due to using a large model and multiple nodes with lengthy instructions.
    In comparison, the Student agent achieves an honorable \textbf{76.6\%} GREADTH score comparable to the Teacher agent but costing only \textbf{\$85/1M} agent turns, a \textbf{factor 300x cheaper}, thanks to shorter instructions and smaller model.
    \item We also evaluate single-node agents (same as the Student) but with different LMs in a 0-shot setting, with no fine-tuning. We achieve GREADTH scores of only 43.2\% with Llama-3.1-405B and 2.0\% with Llama-3.1-8B. This is not particularly surprising given that the single-node Student agent prompt is designed for cost optimization over task success, though it is interesting to observe that the big gap between the two LLMs ($43.2\%-2.0\%=41.2\%\textrm{\ points}$) is entirely closed by the finetuning process (the finetuned Student is 0.8\% points above the Teacher).
    \item Similarly, we evaluate a multi-node agent (same as the Teacher) but with a small LM (Llama-3.1-8B-Instruct) in a 0-shot setting, with no fine-tuning. We observe a large difference of $75.8\%-36.6\%=39.2\%\textrm{ points}$, which mostly confirms that while using the largest models is most crucial in the zero-shot regime, finetuning smaller models can close the gap and generalize across domains.
\end{enumerate}

% \paragraph{RLAIF with GRPO} We experiment with the GRPO~\citep{shao2024deepseekmath} reinforcement learning from human feedback (RLHF) algorithm. We implement a GREADTH critique by prompting a Llama-3.1-405B-Instruct-FP8 to predict whether each agent response satisfies the 6 metrics previously defined (binary predictions). We also consider a 7th metric which is satisfied only if the model output is parsable into the desired output schema and satisfies programmatic constraints (e.g. cannot fill a categorical slot with other values).
% We define the RL reward as a linear combination of the 7 metrics:

% \begin{center}
% \begin{tabular}{lc}
% \toprule
% \textbf{Metric} & \textbf{Weight} \\
% \midrule
% Accurate & 0.4 \\
% Grounded & 1.0 \\
% Transparent1 & 0.2 \\
% Transparent2 & 0.5 \\
% Helpful & 0.4 \\
% NoError & 2.0 \\
% \bottomrule
% \end{tabular}
% \end{center}

\section{Related Work}
\label{sec:related_work}

Autonomous agents aim to tackle complex tasks via self-guided planning and actions in often dynamic environments \citep{wang2024survey}. Several strategies have been introduced to streamline agent's abilities to accomplish their objectives. One tried-and-true strategy to enhance the reasoning capabilities involves meta-reasoning where the agent is instructed to generate reasoning traces and plans prior to arriving at a final decision \citep{DBLP:conf/iclr/YaoZYDSN023}. Another prominent strategy hinges on improving the agent's ability to conduct self-diagnosis \citep{xie2023selfevaluation,kim2023language,shinn2024reflexion,chen2024teaching}. Combining both strategies to find a success path from different reasoning paths and self-evaluating choices has also proven effective \citep{yao2023tree}. TapeAgents offers a holistic solution for seamlessly integrating these paradigms into agent behavior. In particular, \textit{Steps} in TapeAgents represents planning and reasoning traces that guide the agent's actions and \textit{Nodes}  can be used for self-evaluation. Finally, \textit{Tapes} serves as an abstraction for memory that logs agent's internal thought process and interactions with the environment. In this section, we provide a detailed overview of existing agentic frameworks and examine how they compare to TapeAgents.

Among the developer-oriented frameworks most similar to TapeAgents are LangGraph~\citep{Chase_LangGraph_2023} and AutoGen~\citep{wu2023autogen}. In LangGraph, one builds the agent at a very low-level as a concurrent graph-based state machine. AutoGen offers a high-level paradigm to build multi-agent teams. TapeAgents combines the best of these two worlds, as it allows both the low-level control and the implementation of higher-level low-code paradigms like AutoGen.
Neither LangGraph nor AutoGen are designed with agent optimization in mind. 

On the other side of the spectrum are AI frameworks that have recently demonstrated techniques to automatically optimize prompts and tweak other aspects of agent configuration, such as the flow or the assignment of tools to the agents.
Solutions using DSPy~\citep{khattab2023dspycompilingdeclarativelanguage} and TextGrad~\citep{yuksekgonul2024textgradautomaticdifferentiationtext} attain higher performance compared to human prompt engineering.
AgentOptimizer~\citep{zhang2024training} and agent symbolic learning~\citep{zhou2024symboliclearningenablesselfevolving} enable improving agent tools and pipelines. Meta-Agent Search~\citep{hu2024automateddesignagenticsystems} aims to create the best multi-agent architecture. The rich metadata that links the tape with the agent configuration in TapeAgents provides a perfect medium for implementing agent optimization algorithms like the ones from the above works. Notably, DSPy and TextGrad implement control flow in pure Python, which created challenges for resuming the agent from a persistent session state. In TapeAgents, developers can freely use Python within one node, but between-node control flow is handled by adding steps to the tape, which makes the tape perfect for session persistence.

\subsection{Detailed comparison with LangGraph, AutoGen, DSPy}
To show that TapeAgents is uniquely holistic in targeting all stages of LLM agent lifecycle, we present a more detailed side-by-side comparison of TapeAgents with three particularly popular frameworks: LangGraph, AutoGen\footnote{A new AutoGen 0.4 version with major changes came out when we were writing this report. The comparison below is based on a prior version. } and DSPy. We focus on the following seven axis that are particularly helpful to differentiate these frameworks:
\begin{enumerate}
    % https://langchain-ai.github.io/langgraph/how-tos/subgraph/
    \item \textbf{Building from Components while Allowing Finegrained Flow Control} All the frameworks we compare here allow building agentic systems from components, which are called agents in TapeAgents and AutoGen, subgraphs in LangGraph, predictors in DSPy. TapeAgents and LangGraph additionally offer fine-grainted control flow inside each module via nodes and transition between them. To have the same level of control in AutoGen, one must write long non-resumable blocks of Python code.
    \item \textbf{Native Streaming Support}
    LangGraph and TapeAgents natively support response streaming by propagating intermediate event loops through the orchestration loops. Streaming support for AutoGen agents requires substantial change to the agent implementation. 
    \item \textbf{Concurrent LLM Calls} 
    LangGraph state-machine natively supports concurrent node execution, which allows one to make concurrent LLM Calls. At the moment the agent in TapeAgents can only run one node of one agent at a time. We have a plan on how to address this limitation, see Section \ref{sec:future_work} for details. We believe AutoGen also has no intra-agent concurrency at the framework level. DSPy users orchestrate DSPy modules with Python code, hence they can use Python multithreading to run multiple modules concurrently.
    \item \textbf{Resumable State Machine Agents} Developer-friendly frameworks like LangGraph and TapeAgents share the following design pattern: (a) there is a notion of the agent state (tape) that is serializable and that fully determines the agent's behavior (b) one implements an agent by defining how the agent makes the new state from the old state in response to external inputs, such as LLM outputs or API responses. This \textit{state machine agent} pattern should be contrasted with implementing the agent in pure Python code as advocated by DSPy. In the latter case the agent state is entangled in the non-serializable state of the Python interpreter. The advantage of having a well-defined state machine for the agent is that it can resume from frequent state checkpoints, such as intermediate tapes in case of TapeAgents, and it can be stopped at any time. Our understanding is that AutoGen provides less control over resumption and agent execution, e.g. one can not just rerun the next speaker selection or restart the agent from a pre-recorded speaker selection.
    \item \textbf{Log Reuse Across Agents}
    A practitioner that uses TapeAgents can reuse tapes from one agent to evaluate or improve another agent. We show an example of this pattern in our form-filling case-study in \cref{sec:ghreatintro}. There, the tapes could be reused as is, though we believe often minor modifications can suffice for adapting the tape from e.g. a monolithic agent to an agent team. Our understanding is that from the frameworks in comparison only LangGraph can be modified to support this pattern.
    \item \textbf{Structured Logs and Agent Configurations for Data-Driven Agent Optimization}
    In TapeAgents the agent configuration and the tape are highly structured and linked with metadata, helping the implementation of agent optimization algorithms. LangGraph does not currently position its event sequences as a medium to express algorithms. We likewise found the message history and the agent configuration in AutoGen were not designed with agent optimization in mind. On the contrary, the hierarchical structure of DSPy predictors and the traces that DSPy programs enabled implementation of numerous effective algorithms.
    \item \textbf{Making Training Text From Semantic-Level Logs}
    Tape offers a semantic-level representation of the agent session that the agent can convert into low-level training text for LLM finetuning. This is a unique trait of our framework which we believe LangGraph and AutoGen do not share. We believe with some effort a similar pattern could be implemented in DSPy. 
\end{enumerate}

We refer the reader to Table \ref{tab:related_work} for a tabular summary of the above analysis. One can see that TapeAgents uniquely helps practitioners to both develop the agent and optimize it in a data-driven way. 

\subsection{Observability Platforms vs TapeAgents}

Agent observability software such as LangSmith\footnote{\href{https://www.langchain.com/langsmith}{{https://www.langchain.com/langsmith}}}
and Langfuse\footnote{\href{https://github.com/langfuse/langfuse}{https://github.com/langfuse/langfuse}} adds visibility to agent execution.
They allow one to incrementally instrument agent code to track specific components. In TapeAgents, the tape offers complete observability by design, but beyond that it can also be used for point-in-time resumption and agent optimization. 

% \begin{figure}
%     \centering
%     \includegraphics[width=0.5\linewidth]{figures/related_frameworks.pdf}
%     \caption{\textbf{TapeAgents vs Other Frameworks.} TapeAgents stands out in how it supports the practitioner in tasks throughout the development cycle. In this figure we use the stop sign to indicate that major core changes would be required for the framework to be helpful for a task. Warning sign indicates partial support. We indicate limitations of serving support in TapeAgents that we will address in future work, see Section \ref{sec:future_work} for a longer discussion.}
%     \label{fig:related_frameworks}
% \end{figure}

% \ding{51}
% \ding{115}
% \ding{55}
\begin{table}[t]
% \begin{adjustbox}{width=\columnwidth}
    \centering
    \adjustbox{max width=\columnwidth}{
    \begin{tabular}{c|p{3cm}|p{2cm}|p{2cm}|p{2cm}|p{2cm}|p{3cm}|p{2cm}}
    \multicolumn{1}{c|}{} &
    \multicolumn{4}{c|}{\textbf{Development}} &
    \multicolumn{3}{c}{\textbf{Optimization}} \\
    \hline
    \multicolumn{1}{c|}{\textbf{Method}} &
    \multicolumn{1}{p{3cm}|}{\textbf{Building from Components while Allowing Finegrained Flow Control}} & 
    \multicolumn{1}{p{2cm}|}{\textbf{Native Streaming Support}} & 
    \multicolumn{1}{p{1cm}|}{\textbf{Concurrent LLM Calls}} & 
    \multicolumn{1}{p{2cm}|}{\textbf{Resumable State Machine Agents}} & 
    \multicolumn{1}{p{2cm}|}{\textbf{Log Reuse Across Agents}} & 
    \multicolumn{1}{p{3.5cm}|}{\textbf{Structured Logs and Agent Configurations for Data-Driven Agent Optimization}} &
    \multicolumn{1}{p{2cm}}{\textbf{Making Training Text From Semantic-Level Logs}}\\
    \hline
    DSPy%~\cite{khattab2023dspycompilingdeclarativelanguage}
    & \ding{51} & \ding{55}   & \ding{51} & \ding{55}  & \ding{55} & \ding{51} & \ding{115}\\
    LangGraph%~\cite{Chase_LangGraph_2023}
    & \ding{51} & \ding{51}   & \ding{51} & \ding{51}  & \ding{115} & \ding{115} & \ding{55}\\
    AutoGen%~\cite{wu2023autogen}~\cite{zhang2024training}~\citep{dibia2024studio} 
    & \ding{115} & \ding{115} & \ding{55} & \ding{115} & \ding{55} & \ding{115} & \ding{55}\\
    \hline
    TapeAgents (Ours)
    & \ding{51} & \ding{51}   & \ding{55} & \ding{51}  & \ding{51} & \ding{51} & \ding{51}\\
    \end{tabular}
}
    \caption{\textbf{TapeAgents vs Other Frameworks.} TapeAgents stands out in features it offers to  the practitioner to the support them throughout the LLM Agent development cycle. In this figure, we use the cross sign (\ding{55}) to indicate that major core changes would be required for the framework support the feature. Triangle sign (\ding{115}) indicates partial support of a feature, meaning that practitioner would have to do extra effort or accept associated limitations to achieve the respective functionality. Check sign (\ding{51}) indicates that the framework natively supports a feature. TapeAgents's only weakness in this table is the lack of Concurrent LLM Calls, see Section \ref{sec:future_work} for a discuss of how we intend to tackle it.}
    % \caption{\textbf{TapeAgents vs Other Frameworks.} TapeAgents stands out in how it supports the practitioner in tasks throughout the development cycle. In this figure, we use the cross sign (\ding{55}) to indicate that major core changes would be required for the framework to be helpful for a task. Triangle sign (\ding{115}) indicates partial support, meaning th. Check sign (\ding{51}) indicates that the framework is compatible with a task. We indicate limitations of serving support in TapeAgents that we will address in future work, see Section \ref{sec:future_work} for a longer discussion.}
    \label{tab:related_work}
    % \end{adjustbox}
\end{table}

% \begin{table}[h]
% % \begin{adjustbox}{width=\columnwidth}
%     \centering
%     \adjustbox{max width=\columnwidth}{
%         \begin{tabular}{c|cccc}
%         \multicolumn{1}{c|}{\textbf{Method}} &
%         \multicolumn{1}{c|}{\textbf{Build}} &
%         \multicolumn{1}{c|}{\textbf{Debug}} & 
%         \multicolumn{1}{c|}{\textbf{Serve}} & 
%         \multicolumn{1}{c}{\textbf{Optimize}} \\
%         \hline
%         DSPy~\citep{khattab2023dspycompilingdeclarativelanguage} & \ding{55} & \ding{55} & \ding{55} & \ding{51}\\
%         Agents~\citep{zhou2024symboliclearningenablesselfevolving}~\citep{zhou2023agents} & \ding{51} & \ding{55} & \ding{51} & \ding{51} \\
%         LangGraph~\citep{Chase_LangGraph_2023} &\ding{51} & \ding{51} & \ding{51} & \ding{55} \\
%         AutoGen~\citep{wu2023autogen}~\citep{zhang2024training}~\citep{dibia2024studio} & \ding{115} & \ding{115} & \ding{51} & \ding{115} \\
%         \hline
%         TapeAgents (Ours) 
%         % \char"D83D  
%         \char"0001F534
%         & \trafficlight{1} & \ding{51} & \ding{51} & \ding{51}\\
%         \end{tabular}
%     }
%     \caption{\textbf{Agentic Frameworks.} TapeAgents is a unique holistic frameworks that target all stages of the AI Agent lifecycle.}
%     \label{tab:related_work}
%     % \end{adjustbox}
% \end{table}

\section{Discussion and Future Work}
\label{sec:future_work}
We have presented TapeAgents, a holistic framework that targets all stages of the LLM Agent lifecycle. We believe the tape-centered approach of our framework can facilitate responsible deployment and continual improvement of LLM agents. Initial tapes will help debugging and testing at the development stages, historical tapes from production agent sessions will serve as a machine-readable source of evaluation and training data. Red-teaming algorithms can use historical tapes as seed data for testing the agent on potential attacks or business-critical dangerous situations. The practitioner can also use historical tapes to seed the simulation that they use for testing the agent. These are but a few benefits that TapeAgents can bring to practitioners.

\subsection{Immediate Next Steps}
TapeAgents is still in early development stages. A key next step for TapeAgents is adding coroutine implementations for the agent loop and for agent-environment orchestration. This will enable both running many agent-environment loops in parallel on their respective tapes and running members of the same agent team on their shared tape. The latter will require changes in the tape view computation to ignore steps of the agents running in parallel, but we believe that the main framework concepts that we introduced in this paper will stay the same. 

On the optimization front, we will soon release an online Reinforcement Learning (RL) trainer for TapeAgents, which will improve the assistant agent using the rewards that the annotator agent computes. Another welcome optimizer addition would be implementing in TapeAgents a text-based feedback-propagation algorithm like the one in TextGrad. Tapes steps are a perfect medium to attach feedback to.

\subsection{Agent as an Optimizable Workflow}
Stepping back from the immediate future plans, we believe it is worth reflecting on what should be called an LLM Agent and what should be called ``just'' a program, a workflow or software. In the TapeAgents context, this philosophical question is what the developer asks themselves when they build an agent---whether they should implement as agent nodes, and what should go in the application that uses the agent. Our current recommendation is that one should treat and implement the parts of the system that they intend to optimize with data-driven algorithms as LLM Agents. Frameworks provide the structure that the algorithms require to identify the issue, propose a change, and test the change's outcome. In TapeAgents, this process is particularly clear: the algorithm will identify an issue in the tape, attribute it to a root cause step, propose a change to the agent configuration and test this change by resuming the agent from intermediate points in the tape. Thus the gains from algorithmic improvement will compensate for the overhead of respecting the TapeAgents engineering constraints. To sum up, our recommendation is to implement \emph{optimizable workflows} as LLM Agents and use other appropriate tools for the software that will not be subject to data-driven improvement. 

\subsection{Synthetic Data Generation with Worlds of TapeAgents}
In addition to helping practitioners with their solution-specific challenges, we envision synthetic data generation as another application area where TapeAgents can make an impact. A key trend in the data-making trade is building modular pipelines with many agent-like modules, such as prompt-generating workflows~\citep{mitra2024agentinstruct}, judges~\citep{bai2022constitutionalaiharmlessnessai}, meta-judges~\citep{wu2024meta}, process supervisors~\citep{lightman2023let, uesato2022solvingmathwordproblems}, annotator augmented with tools~\citep{wei2024longformfactualitylargelanguage} among other examples. We believe TapeAgents is a great foundation for the continual improvement of such multi-agent pipelines with human feedback, as implementing all pipeline modules as TapeAgents immediately makes them optimizable. 
% TODO: more inspiring stuff here

% \begin{table}[t]
% \centering
% \begin{tabular}{lc}
% \toprule
% \textbf{Approach} & Win Rate \\
% \midrule
% Distilled Student vs Teacher & 50 \\
% Rej. Sampling vs Teacher & 50 \\
% Reinforce with Critique vs Teacher & 50 \\
% \bottomrule
% \end{tabular}
% \caption{\textbf{Experience evaluation}. }
% \label{tab:factscore}
% \end{table}

%%% results:
% evaluation of critique?
% win rates over models that have GREADTH eval from critique

%%% analysis:
% table with user behavior distributions - breakdown 
% distribution of predicted steps
\FloatBarrier

\section*{Acknowledgements}
\label{sec:acknowledgements}
We would like to thank Nicolas Chapados, Chris Manning, Chris Pal, Sebastien Paquet, Siva Reddy, Arkil Patel and David Vazquez for their feedback and ideas.
\FloatBarrier

\bibliographystyle{apalike}
\bibliography{nowai,bibliography}

\begin{thebibliography}{}

\bibitem[Bai et~al., 2022]{bai2022constitutionalaiharmlessnessai}
Bai, Y., Kadavath, S., Kundu, S., Askell, A., Kernion, J., Jones, A., Chen, A., Goldie, A., Mirhoseini, A., McKinnon, C., Chen, C., Olsson, C., Olah, C., Hernandez, D., Drain, D., Ganguli, D., Li, D., Tran-Johnson, E., Perez, E., Kerr, J., Mueller, J., Ladish, J., Landau, J., Ndousse, K., Lukosuite, K., Lovitt, L., Sellitto, M., Elhage, N., Schiefer, N., Mercado, N., DasSarma, N., Lasenby, R., Larson, R., Ringer, S., Johnston, S., Kravec, S., Showk, S.~E., Fort, S., Lanham, T., Telleen-Lawton, T., Conerly, T., Henighan, T., Hume, T., Bowman, S.~R., Hatfield-Dodds, Z., Mann, B., Amodei, D., Joseph, N., McCandlish, S., Brown, T., and Kaplan, J. (2022).
\newblock Constitutional ai: Harmlessness from ai feedback.

\bibitem[Budzianowski et~al., 2018]{budzianowski2018large}
Budzianowski, P., Wen, T.-H., Tseng, B.-H., Casanueva, I., Stefan, U., Osman, R., and Ga{\v{s}}i\'c, M. (2018).
\newblock Multiwoz - a large-scale multi-domain wizard-of-oz dataset for task-oriented dialogue modelling.
\newblock In {\em Proceedings of the 2018 Conference on Empirical Methods in Natural Language Processing (EMNLP)}.

\bibitem[Chase, 2022]{Chase_LangChain_2022}
Chase, H. (2022).
\newblock {LangChain}.

\bibitem[Chase, 2023]{Chase_LangGraph_2023}
Chase, H. (2023).
\newblock {LangGraph}.

\bibitem[Chen et~al., 2024]{chen2024teaching}
Chen, X., Lin, M., Sch{\"a}rli, N., and Zhou, D. (2024).
\newblock Teaching large language models to self-debug.
\newblock In {\em The Twelfth International Conference on Learning Representations}.

\bibitem[Drouin et~al., 2024]{drouin2024workarena}
Drouin, A., Gasse, M., Caccia, M., Laradji, I.~H., Verme, M.~D., Marty, T., Vazquez, D., Chapados, N., and Lacoste, A. (2024).
\newblock Workarena: How capable are web agents at solving common knowledge work tasks?
\newblock In {\em Forty-first International Conference on Machine Learning}.

\bibitem[Fourney et~al., 2024]{fourney2024magentic}
Fourney, A., Bansal, G., Mozannar, H., Tan, C., Salinas, E., Niedtner, F., Proebsting, G., Bassman, G., Gerrits, J., Alber, J., et~al. (2024).
\newblock Magentic-one: A generalist multi-agent system for solving complex tasks.
\newblock {\em arXiv preprint arXiv:2411.04468}.

\bibitem[Gugger et~al., 2022]{accelerate}
Gugger, S., Debut, L., Wolf, T., Schmid, P., Mueller, Z., Mangrulkar, S., Sun, M., and Bossan, B. (2022).
\newblock Accelerate: Training and inference at scale made simple, efficient and adaptable.
\newblock \url{https://github.com/huggingface/accelerate}.

\bibitem[Hu et~al., 2021]{hu2021LoRA}
Hu, E.~J., Shen, Y., Wallis, P., Allen-Zhu, Z., Li, Y., Wang, S., Wang, L., and Chen, W. (2021).
\newblock Lora: Low-rank adaptation of large language models.
\newblock {\em arXiv preprint arXiv: 2106.09685}.

\bibitem[Hu et~al., 2024]{hu2024automateddesignagenticsystems}
Hu, S., Lu, C., and Clune, J. (2024).
\newblock Automated design of agentic systems.

\bibitem[Khattab et~al., 2023a]{khattab2023dspy}
Khattab, O., Singhvi, A., Maheshwari, P., Zhang, Z., Santhanam, K., Vardhamanan, S., Haq, S., Sharma, A., Joshi, T.~T., Moazam, H., Miller, H., Zaharia, M., and Potts, C. (2023a).
\newblock {DSPy}: Compiling declarative language model calls into self-improving pipelines.
\newblock {\em arXiv preprint arXiv:2310.03714}.

\bibitem[Khattab et~al., 2023b]{khattab2023dspycompilingdeclarativelanguage}
Khattab, O., Singhvi, A., Maheshwari, P., Zhang, Z., Santhanam, K., Vardhamanan, S., Haq, S., Sharma, A., Joshi, T.~T., Moazam, H., Miller, H., Zaharia, M., and Potts, C. (2023b).
\newblock {DSPy}: Compiling declarative language model calls into self-improving pipelines.

\bibitem[Kim et~al., 2023]{kim2023language}
Kim, G., Baldi, P., and McAleer, S.~M. (2023).
\newblock Language models can solve computer tasks.
\newblock In {\em Thirty-seventh Conference on Neural Information Processing Systems}.

\bibitem[Lightman et~al., 2023]{lightman2023let}
Lightman, H., Kosaraju, V., Burda, Y., Edwards, H., Baker, B., Lee, T., Leike, J., Schulman, J., Sutskever, I., and Cobbe, K. (2023).
\newblock Let's verify step by step.
\newblock {\em arXiv preprint arXiv:2305.20050}.

\bibitem[Loshchilov and Hutter, 2017]{loshchilov2017decoupled}
Loshchilov, I. and Hutter, F. (2017).
\newblock Decoupled weight decay regularization.
\newblock {\em International Conference on Learning Representations}.

\bibitem[Mialon et~al., 2024]{mialon2024gaia}
Mialon, G., Fourrier, C., Wolf, T., LeCun, Y., and Scialom, T. (2024).
\newblock {GAIA}: a benchmark for general {AI} assistants.
\newblock In {\em The Twelfth International Conference on Learning Representations}.

\bibitem[Mitra et~al., 2024]{mitra2024agentinstruct}
Mitra, A., Corro, L.~D., Zheng, G., Mahajan, S., Rouhana, D., Codas, A., Lu, Y., ge~Chen, W., Vrousgos, O., Rosset, C., Silva, F., Khanpour, H., Lara, Y., and Awadallah, A. (2024).
\newblock Agentinstruct: Toward generative teaching with agentic flows.

\bibitem[Pryzant et~al., 2023]{pryzant-etal-2023-automatic}
Pryzant, R., Iter, D., Li, J., Lee, Y., Zhu, C., and Zeng, M. (2023).
\newblock Automatic prompt optimization with {``}gradient descent{''} and beam search.
\newblock In {\em Proceedings of the 2023 Conference on Empirical Methods in Natural Language Processing}, pages 7957--7968, Singapore. Association for Computational Linguistics.

\bibitem[Rasley et~al., 2020]{Rasley2020DeepSpeedSO}
Rasley, J., Rajbhandari, S., Ruwase, O., and He, Y. (2020).
\newblock Deepspeed: System optimizations enable training deep learning models with over 100 billion parameters.
\newblock {\em Proceedings of the 26th ACM SIGKDD International Conference on Knowledge Discovery \& Data Mining}.

\bibitem[Rastogi et~al., 2020]{rastogi2020towards}
Rastogi, A., Zang, X., Sunkara, S., Gupta, R., and Khaitan, P. (2020).
\newblock Towards scalable multi-domain conversational agents: The schema-guided dialogue dataset.
\newblock In {\em Proceedings of the AAAI Conference on Artificial Intelligence}, volume~34, pages 8689--8696.

\bibitem[Shinn et~al., 2024]{shinn2024reflexion}
Shinn, N., Cassano, F., Gopinath, A., Narasimhan, K., and Yao, S. (2024).
\newblock Reflexion: Language agents with verbal reinforcement learning.
\newblock In {\em Advances in Neural Information Processing Systems}, volume~36.

\bibitem[Uesato et~al., 2022]{uesato2022solvingmathwordproblems}
Uesato, J., Kushman, N., Kumar, R., Song, F., Siegel, N., Wang, L., Creswell, A., Irving, G., and Higgins, I. (2022).
\newblock Solving math word problems with process- and outcome-based feedback.

\bibitem[Wang et~al., 2024]{wang2024survey}
Wang, L., Ma, C., Feng, X., Zhang, Z., Yang, H., Zhang, J., Chen, Z., Tang, J., Chen, X., Lin, Y., Zhao, W.~X., Wei, Z., and Wen, J. (2024).
\newblock A survey on large language model based autonomous agents.
\newblock {\em Frontiers of Computer Science}, 18(6):186345.

\bibitem[Wei et~al., 2024]{wei2024longformfactualitylargelanguage}
Wei, J., Yang, C., Song, X., Lu, Y., Hu, N., Huang, J., Tran, D., Peng, D., Liu, R., Huang, D., Du, C., and Le, Q.~V. (2024).
\newblock Long-form factuality in large language models.

\bibitem[Wu et~al., 2024a]{wu2023autogen}
Wu, Q., Bansal, G., Zhang, J., Wu, Y., Li, B., Zhu, E., Jiang, L., Zhang, X., Zhang, S., Liu, J., Awadallah, A.~H., White, R.~W., Burger, D., and Wang, C. (2024a).
\newblock Autogen: Enabling next-gen llm applications via multi-agent conversation framework.
\newblock In {\em COLM}.

\bibitem[Wu et~al., 2024b]{wu2024meta}
Wu, T., Yuan, W., Golovneva, O., Xu, J., Tian, Y., Jiao, J., Weston, J., and Sukhbaatar, S. (2024b).
\newblock Meta-rewarding language models: Self-improving alignment with {LLM}-as-a-meta-judge.
\newblock {\em arXiv preprint arXiv:2407.19594}.

\bibitem[Xie et~al., 2023]{xie2023selfevaluation}
Xie, Y., Kawaguchi, K., Zhao, Y., Zhao, X., Kan, M.-Y., He, J., and Xie, Q. (2023).
\newblock Self-evaluation guided beam search for reasoning.
\newblock In {\em Thirty-seventh Conference on Neural Information Processing Systems}.

\bibitem[Yao et~al., 2023a]{yao2023tree}
Yao, S., Yu, D., Zhao, J., Shafran, I., Griffiths, T.~L., Cao, Y., and Narasimhan, K.~R. (2023a).
\newblock Tree of thoughts: Deliberate problem solving with large language models.
\newblock In {\em Thirty-seventh Conference on Neural Information Processing Systems}.

\bibitem[Yao et~al., 2023b]{yao2023reactsynergizingreasoningacting}
Yao, S., Zhao, J., Yu, D., Du, N., Shafran, I., Narasimhan, K., and Cao, Y. (2023b).
\newblock {ReAct}: Synergizing reasoning and acting in language models.

\bibitem[Yao et~al., 2023c]{DBLP:conf/iclr/YaoZYDSN023}
Yao, S., Zhao, J., Yu, D., Du, N., Shafran, I., Narasimhan, K.~R., and Cao, Y. (2023c).
\newblock {ReAct}: Synergizing reasoning and acting in language models.
\newblock In {\em The Eleventh International Conference on Learning Representations, {ICLR} 2023, Kigali, Rwanda, May 1-5, 2023}. OpenReview.net.

\bibitem[Yuksekgonul et~al., 2024]{yuksekgonul2024textgradautomaticdifferentiationtext}
Yuksekgonul, M., Bianchi, F., Boen, J., Liu, S., Huang, Z., Guestrin, C., and Zou, J. (2024).
\newblock Textgrad: Automatic "differentiation" via text.

\bibitem[Zhang et~al., 2024]{zhang2024training}
Zhang, S., Zhang, J., Liu, J., Song, L., Wang, C., Krishna, R., and Wu, Q. (2024).
\newblock Training language model agents without modifying language models.
\newblock {\em ICML'24}.

\bibitem[Zhou et~al., 2023a]{zhou2023agents}
Zhou, W., Jiang, Y.~E., Li, L., Wu, J., Wang, T., Qiu, S., Zhang, J., Chen, J., Wu, R., Wang, S., Zhu, S., Chen, J., Zhang, W., Tang, X., Zhang, N., Chen, H., Cui, P., and Sachan, M. (2023a).
\newblock Agents: An open-source framework for autonomous language agents.

\bibitem[Zhou et~al., 2024]{zhou2024symboliclearningenablesselfevolving}
Zhou, W., Ou, Y., Ding, S., Li, L., Wu, J., Wang, T., Chen, J., Wang, S., Xu, X., Zhang, N., Chen, H., and Jiang, Y.~E. (2024).
\newblock Symbolic learning enables self-evolving agents.

\bibitem[Zhou et~al., 2023b]{zhou2023large}
Zhou, Y., Muresanu, A.~I., Han, Z., Paster, K., Pitis, S., Chan, H., and Ba, J. (2023b).
\newblock Large language models are human-level prompt engineers.
\newblock In {\em The Eleventh International Conference on Learning Representations}.

\end{thebibliography}

\FloatBarrier
\clearpage

\appendix
\label{sec:appendix}

\section{Multi-Agent Code Example}
\label{appendix:multi_agent_code}

\begin{lstlisting}
import json
from tapeagents.agent import Agent, Node
from tapeagents.core import Prompt, SetNextNode, Call, Respond
from tapeagents.dialog_tape import DialogTape, AssistantThought, ToolCalls, UserStep, AssistantStep
from tapeagents.llms import LLMStream, LiteLLM
from tapeagents.prompting import view_to_messages
from tapeagents.orchestrator import main_loop
from tapeagents.tools.simple_browser import SimpleTextBrowser
from tapeagents.tools.stock import get_stock_data, get_stock_ticker
from tapeagents.environment import ToolEnvironment
# to visualize Tapes in Python notebooks
from tapeagents.rendering import render_tape_with_prompts
from tapeagents.renderers.camera_ready_renderer import CameraReadyRenderer
from IPython.display import HTML, clear_output

######### Search Agent #########

# define environment of search agent
browser = SimpleTextBrowser()
search_agent_env = ToolEnvironment([
  browser.get_search_results, browser.get_page, browser.get_next_page
])

# define the main node of the search agent
class SearchAgentMainNode(Node):
  def make_prompt(self, agent, tape: DialogTape) -> Prompt:
    view = agent.compute_view(tape)
    search_system_message = {
      "role": "system",
      "content": """Use at most 5 tool calls to search the request info on on the web."""
    }
    return Prompt(
      messages=[search_system_message] + view_to_messages(view.top, agent), 
      tools=search_agent_env.get_tool_schema_dicts()
    )

  def generate_steps(self, agent, tape, llm_stream: LLMStream):
    m = llm_stream.get_message()
    if m.content:
      # if the LLM responds, yield Respond(..) as your last step. This special step tells the orchestrator that this agent is done. The next active agent will be the one that `Call`ed it (the Analyst Agent in this case).
      yield Respond(content=m.content)
    elif m.tool_calls:
      # while the LLM suggests tool calls, yield them as action steps
      yield ToolCalls.from_llm_output(m)
      yield SetNextNode(next_node=0)  # set yourself as next node for when you get called again
    else:
      raise ValueError()

# create the search agent
search_agent = Agent.create(name="search_agent", llms=LiteLLM(model_name="gpt-4o"), nodes=[SearchAgentMainNode()])
 Agent #########

# define environment of analyst agent

# We will use the tool choice mechanism to let the main agent call its search specialist agent.
# To this end, we create a mock tool that represents calling the search agent.
def call_search_agent(query: str):
    """Use this tool to ask a fellow AI agent to search for information on the web."""
    pass

main_agent_env = ToolEnvironment([
  get_stock_ticker, get_stock_data, call_search_agent
])

today = "2024-09-17"  # fixed date for reproducible tests
system_message = {"role": "system", "content": f"""
You will help the user to learn about financials of companies.  For general user queries, include some info about stock price changes during the last year, as well as some general information on the company. Today is {today}.
"""}

# define the two nodes of the main agent
class PlanNode(Node):
  def make_prompt(self, agent, tape) -> Prompt:
    view = agent.compute_view(tape)
    guidance_message = {
      "role": "user",
      "content": """Write a natural language plan on how to use tools help the user. Output a list of numbered items, like 1., 2., 3., etc."""
    }
    return Prompt(
      messages=[system_message] + view_to_messages(view.top, agent) + [guidance_message],
      tools=main_agent_env.get_tool_schema_dicts(),
    )

  def generate_steps(self, agent, dialog, llm_stream: LLMStream):
    # the PlanNode should only yield thoughts based on the llm output
    if content := llm_stream.get_message().content:
      yield AssistantThought(content=content)
    else:
      raise ValueError()

class ActNode(Node):
  def make_prompt(self, agent, tape: DialogTape) -> Prompt:
    view = agent.compute_view(tape)
    guidance_message = {
      "role": "user",
      "content": """Follow the plan you created to earlier. When you are done, respond to the user."""
    }
    return Prompt(
      messages=[system_message] + view_to_messages(view.top, agent) + [guidance_message],
      tools=main_agent_env.get_tool_schema_dicts(),
    )

  def generate_steps(self, agent, dialog, llm_stream: LLMStream):
    m = llm_stream.get_message()
    if m.content:
      # show the llm output as your response and give back control to plan node
      yield SetNextNode(next_node=0)  # set next node to Plan node
      yield AssistantStep(content=m.content)  # show the llm output
    elif m.tool_calls:
      yield SetNextNode(next_node=1)  # set next node to Act node (self) until we get tool_calls
      # only keep the tool calls before the call to another agent
      agent_call = None
      for i, tc in enumerate(m.tool_calls):
        if tc.function.name == "call_search_agent":
          agent_call = tc
          m.tool_calls = m.tool_calls[:i]
          break
      # either produce the ToolCalls action OR call another agent
      if m.tool_calls:
        yield ToolCalls.from_llm_output(m)
      else:
        assert agent_call and agent_call.function.name == "call_search_agent"
        yield Call(
          agent_name="search_agent",
          content=json.loads(agent_call.function.arguments)["query"]
        )
    else:
      raise ValueError()

# define the main (root) agent
multi_agent_analyst = Agent.create(name="analyst", subagents=[search_agent.clone()], llms=LiteLLM(model_name="gpt-4o"), nodes=[PlanNode(), ActNode()])

# define the starting point: a DialogTape with only 1 UserStep
start_tape = DialogTape(steps=[UserStep(content="Tell me about Vulcan in 3 sentences")])

# define the whole environment as the combination of all agents & subagents environments
whole_env = ToolEnvironment([
  get_stock_ticker, get_stock_data, browser.get_search_results, browser.get_page, browser.get_next_page
])

# Main loop executing the `multi_agent_analyst` on the `start_tape` in the `whole_env` environment.
for event in main_loop(multi_agent_analyst, start_tape, whole_env):
  # `event`s are all types of steps yielded by the `multi_agent_analyst` when running on the `start_tape` in the `whole_env` environment.
  if new_tape := event.agent_tape or event.env_tape:
    # show a fresh render every time when the environment finishes reacting with new actions
    clear_output()
    display(HTML(render_tape_with_prompts(new_tape, CameraReadyRenderer())))

\end{lstlisting}

\newpage

\section{Agentic RAG Code Examples}
\label{app:add-demos}
\begin{lstlisting}[language=Python, caption=add\_demos function]
def add_demos(agent: Agent, tapes: list[Tape], max_n_demos: int, seed: int = 1):
    """Extract demos for function templates from the given tapes.
    When there are too many demos, select random ones.
    """
    demos = {template_name: [] for template_name in agent.templates}
    for tape in tapes:
        for node, index in agent.get_node_runs(tape):
            if isinstance(node, LLMFunctionNode):
                demos[node.template_name].append(node.extract_demo(agent, tape, index))
    rng = random.Random(seed)
    agent_copy = agent.model_copy(deep=True)
    for template_name, template in agent_copy.templates.items():
        k = min(max_n_demos, len(demos[template_name]))
        template.demos = rng.sample(demos[template_name], k)
    return agent_copy
\end{lstlisting}

\label{app:llm-function-templates}
\begin{lstlisting}[language=Python, caption=LLMFunction templates]
# Create a template for answering questions
def make_answer_template() -> LLMFunctionTemplate:
    return LLMFunctionTemplate(
        desc="Answer questions with short factoid answers.",
        inputs=[
            ContextInput(name="context", desc="may contain relevant facts", separator="\n"),
            # The question to be answered
            Input(name="question"),
        ],
        outputs=[
            # Rationale for the answer
            RationaleOutput.for_output("answer"),
            # The actual answer
            AssistantOutput(name="answer", desc="often between 1 and 5 words")
        ]
    )        

# Create a template for generating search queries with a retrieval tool
def make_query_template() -> LLMFunctionTemplate:
    return LLMFunctionTemplate(
        desc="Write a simple search query that will help answer a complex question.",
        inputs=[
            ContextInput(name="context", desc="may contain relevant facts", separator="\n"),
            # The question for which a search query is needed
            Input(name="question"),
        ],
        outputs=[
            # Rationale for the generated query
            RationaleOutput.for_output("query"),
            # The generated search query, to be used with a retrieval tool
            ToolCallOutput(name="query", tool_name="retrieve", arg_name="query")
        ]
    )
\end{lstlisting}

\newpage

\section{Agent Tree and Tape}
\label{appendix:tape}

\begin{figure}[h!]
  \centering
  \begin{minipage}[t]{0.14\textwidth}
    \vspace{0pt}  % Ensures top alignment
    \includegraphics[width=\linewidth, clip]{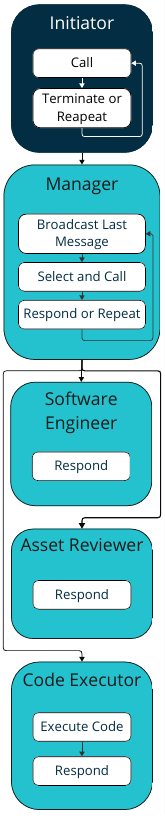}
  \end{minipage}%
  \hfill
  \begin{minipage}[t]{0.85\textwidth}
    \vspace{0pt}  % Ensures top alignment
    \includegraphics[width=\linewidth, clip]{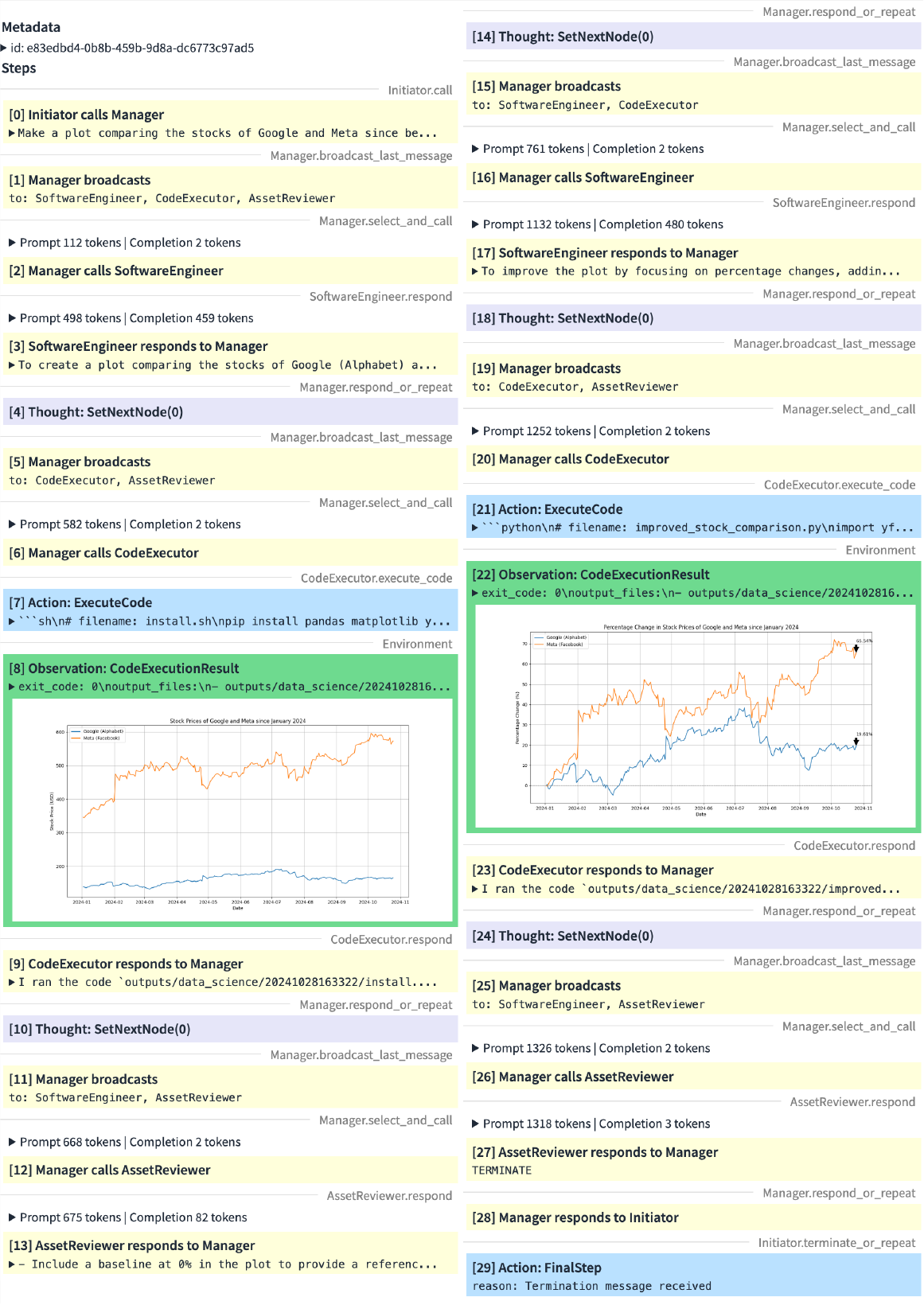}
  \end{minipage}
  \caption{\textit{A multi-agent} tree configuration showing nodes (left) and a tape resulting from their collaboration (right) with color-coded steps: yellow for external agent thoughts (enabling collaboration), purple for internal agent thoughts, blue for actions, and green for observations. The step's author is indicated in grey using the ``Agent.node'' format.}
  \label{fig:multi-agent}
\end{figure}

\section{Tape Tools}
\label{appendix:tools}

\begin{figure}[h!]
  \centering
  \includegraphics[width=1\linewidth,clip]{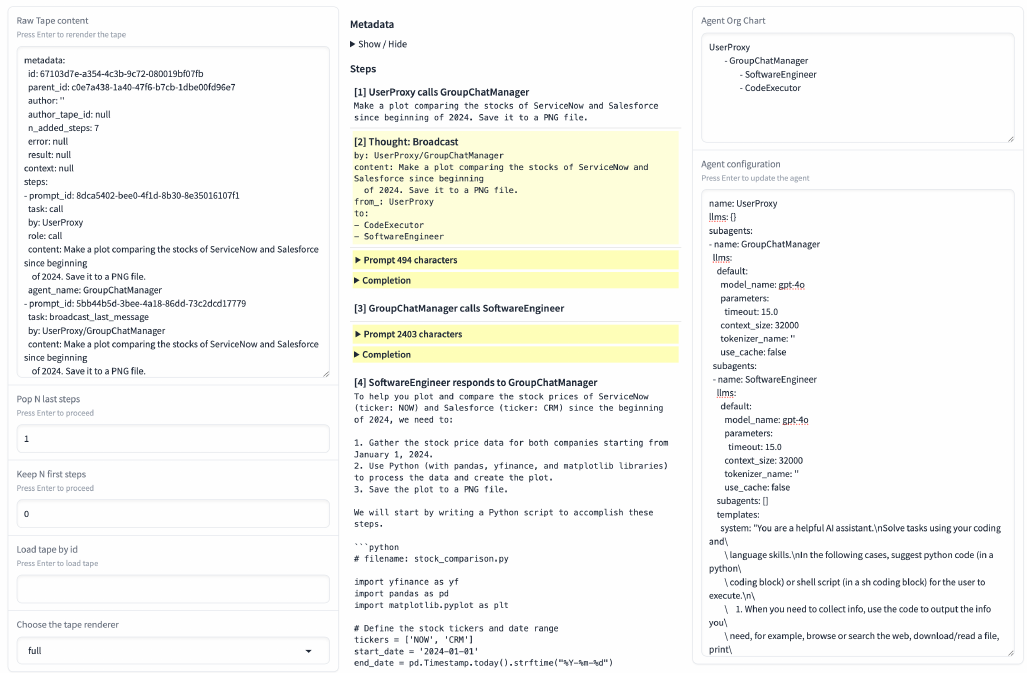}
  \caption{\textbf{TapeAgents Studio}: Application to help AI Admin to edit Tape, resume and debug Agentic Systems}
  \label{fig:studio}
\end{figure}

\newpage

\begin{figure}[h!]
  \centering
  \includegraphics[width=1\linewidth,trim={0cm 6.4cm 0cm 0cm},clip]{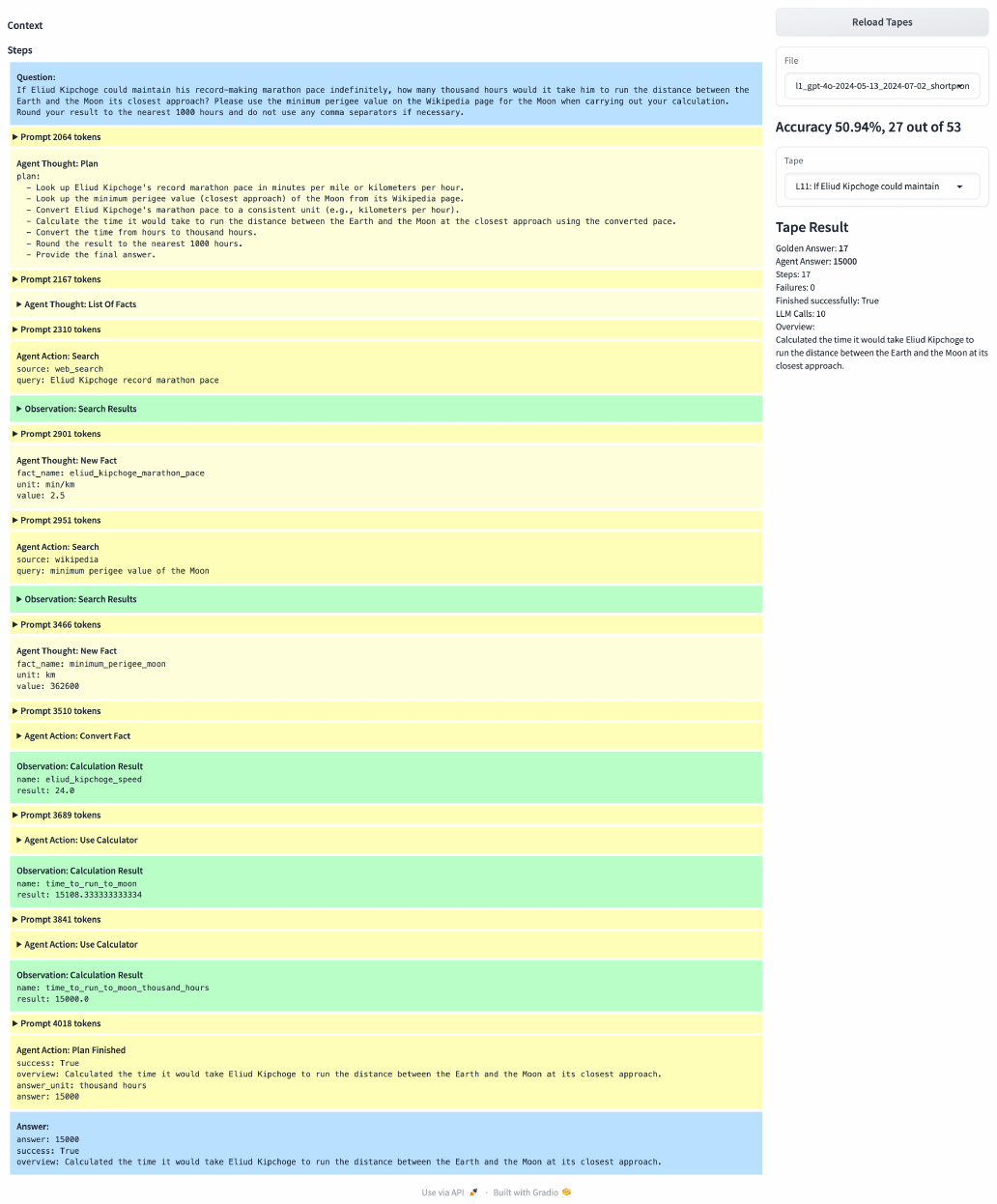}
  \caption{\textbf{Tape Browser}: Application to inspect a batch of tapes result. This tape is a GAIA task where the Agent did not provide the right answer. No step failed during the session.}
  \label{fig:tape-browser}
\end{figure}

\newpage

\begin{figure}[h!]
  \centering
  \includegraphics[width=1\linewidth,clip]{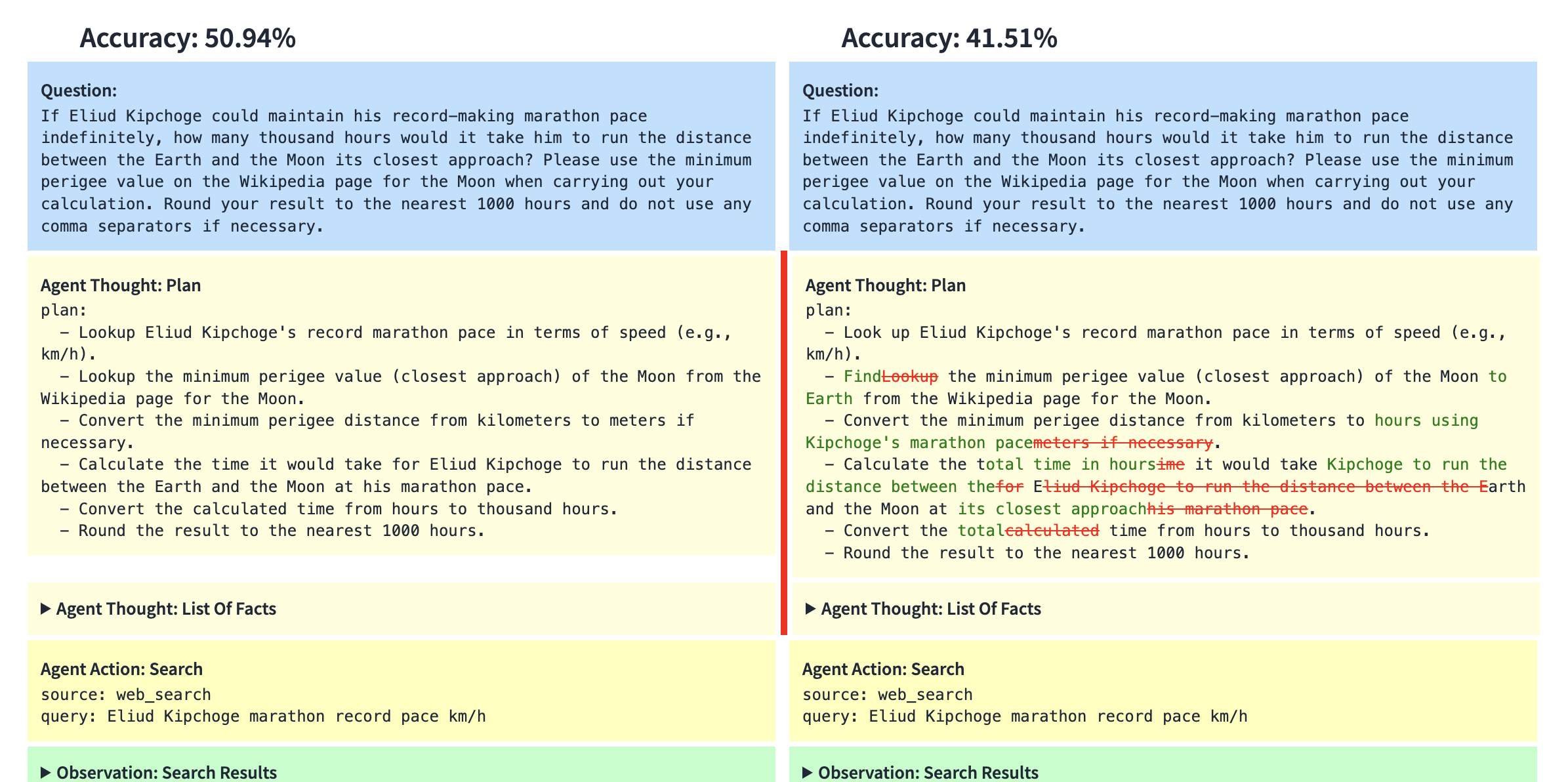}
  \caption{\textbf{Tape Diff}: Application to compare two batches of tape.}
  \label{fig:tape-diff}
\end{figure}

\newpage

\section{GREADTH Form Filler}

\subsection{Virtual Companies Prompts}
\label{appendix:synthetic_companies}

To generate synthetic environments/companies with 10 request forms available in each we use a three-step prompting method with Llama-3-70B.

\begin{figure}[h!]
    \centering
    \includegraphics[width=1\linewidth]{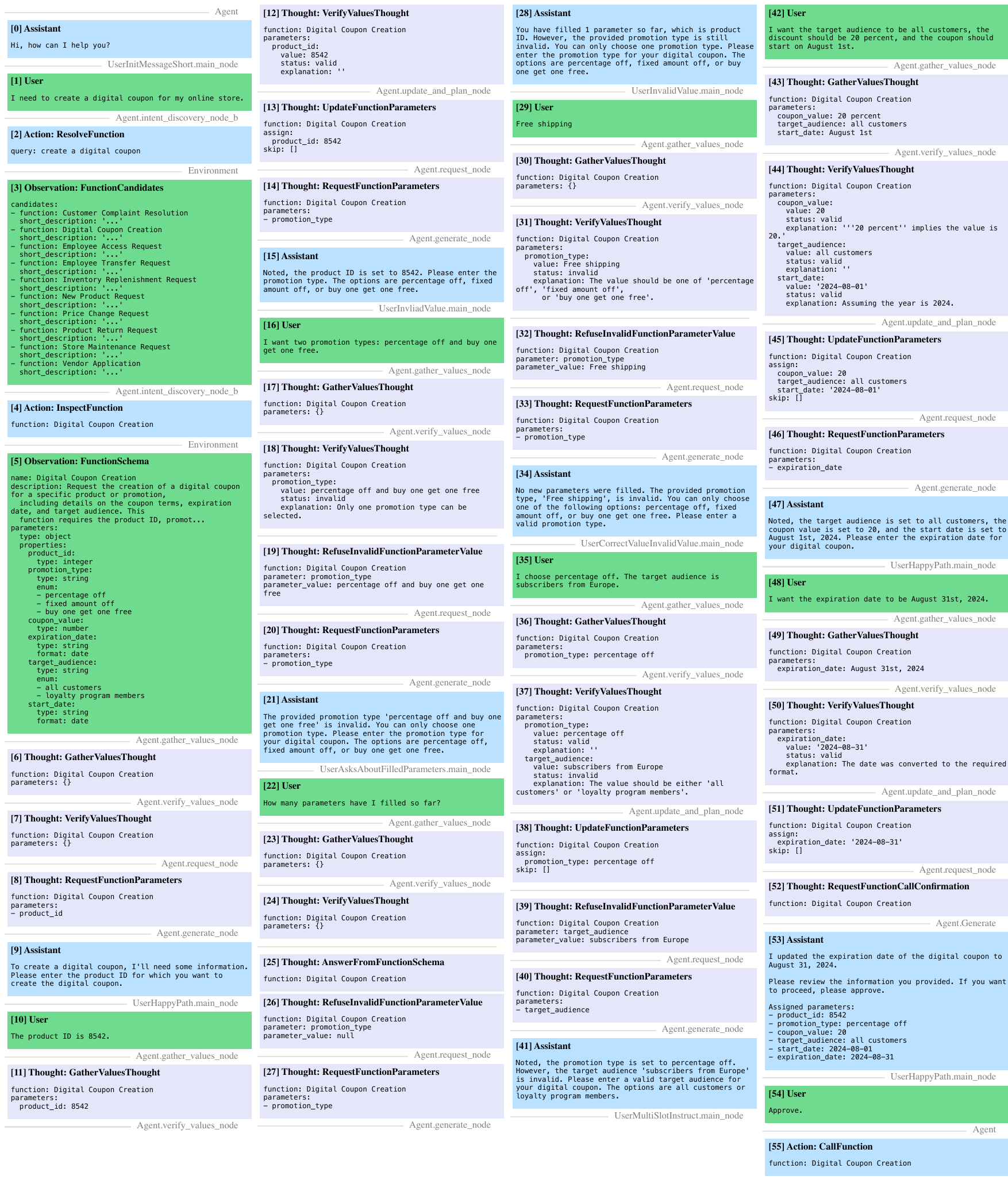}
    \caption{\textbf{Sample Tape between the Teacher agent and User agents}. The user is requesting the creation of a digital coupon for their online store. Steps are color-coded: purple for internal agent thoughts, blue for actions, and green for observations. The step's author is indicated in grey using the ``Agent.node'' format. User agent names are described in Appendix~\ref{appendix:user_behaviors}.}
    \label{fig:teacher_tape}
\end{figure}

\begin{figure}[h!]
    \centering
    \includegraphics[width=1\linewidth]{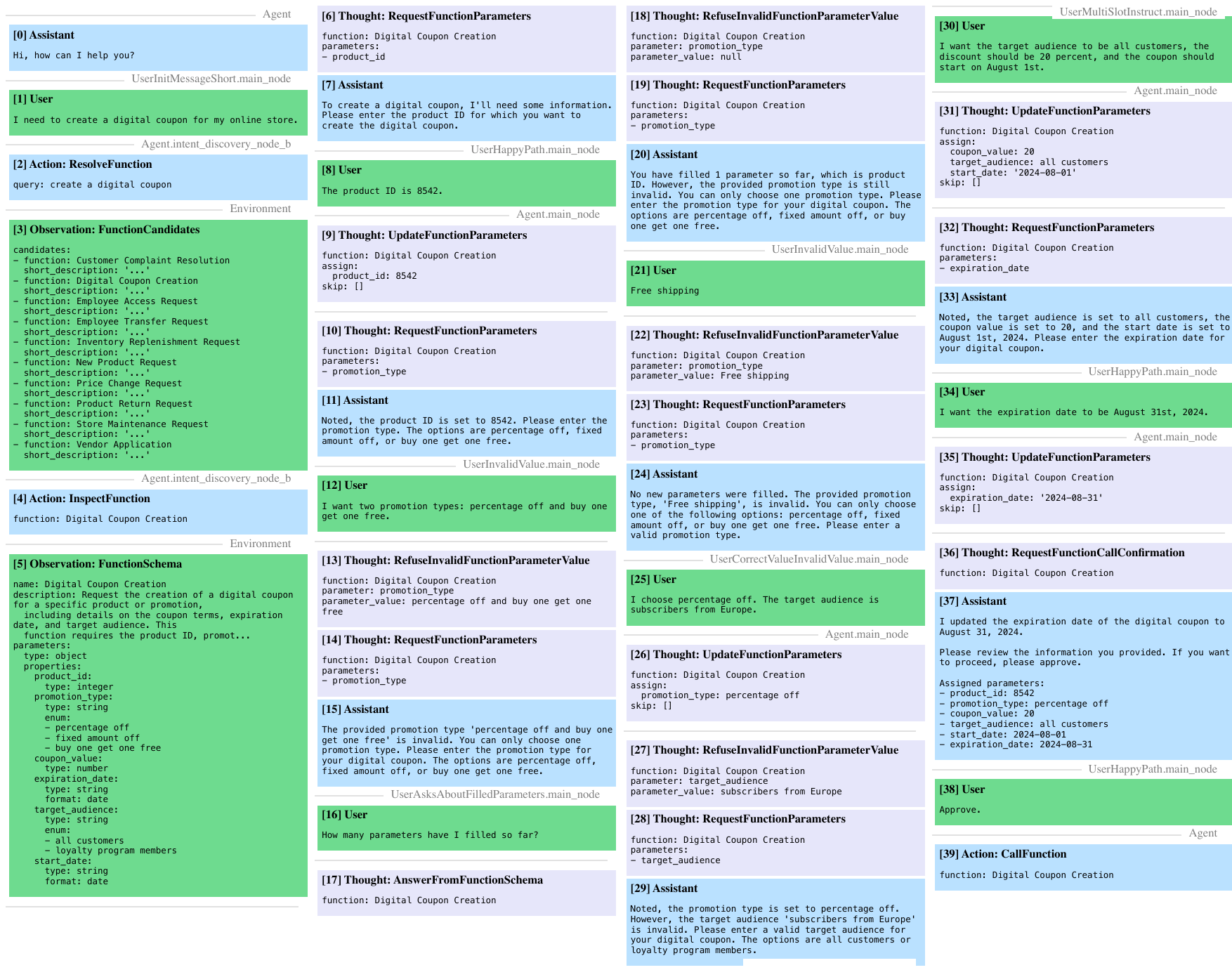}
    \caption{\textbf{Sample tape between the Student agent and User agents}. The user is requesting the creation of a digital coupon for their online store. Steps are color-coded: purple for internal agent thoughts, blue for actions, and green for observations. The step's author is indicated in grey using the ``Agent.node'' format. User agent names are described in Appendix~\ref{appendix:user_behaviors}.}
    \label{fig:student_tape}
\end{figure}

\begin{enumerate}
    \item The first step is to generate a description of a fake company. For this, we use the following prompt:
\begin{verbatim}
messages = [
  {
    "role": "system",
    "content": "You are a helpful assistant."
  },
  {
    "role": "user",
    "content": f"Give me a description of {real_name} but replace all occurrences of
      `{real_name}` by `{fake_name}`."
    }
]
\end{verbatim}
with \texttt{real\_name} and \texttt{fake\_name} being set to different company names (e.g. ``Starbucks'' and ``CoffeeCorp'' respectively). This step produces a \texttt{DESCRIPTION} variable that will be used in the next step.

\item The second step is to generate a list of 10 request forms for each fake company. For now, we only ask the model to generate a name and a description for each form with the following system and user prompts:
\begin{verbatim}
SYSTEM_MESSAGE = """
  You are a helpful enterprise assistant who is very well integrated into the
  internal system of [ENTERPRISE_NAME]. You have access to the database of
  [ENTERPRISE_NAME], which contains REQUEST_FORMS. REQUEST_FORMS can be used
  by employees and clients to submit requests and trigger automations.
  REQUEST_FORMS have the following data format:
  ```python
  class FunctionSchema(BaseModel):
    name: FunctionName
    description: str = Field(description="The description of the function.")
    parameters: JsonSchema = Field(default=None, description="The JSON schema
      of the function's parameters.")
    return_value: JsonSchema = Field(default=None, description="The JSON
      schema of the function's return value.")
  ```
"""
messages = [
  {"role": "system", "content": SYSTEM_MESSAGE},
  {"role": "user", "content": f"""
    [ENTERPRISE_NAME] is '{fake_name}'.
    {DESCRIPTION}
    Give me the name and description of 10 REQUEST_FORMS that are often used
    at {fake_name}. Make sure the REQUEST_FORMS are specific to {fake_name}
    use cases. Format the output as a JSON list of JSON dictionaries with
    'name' and 'description' keys. Make sure the output is JSON parsable.
  """},
]
\end{verbatim}
with \texttt{fake\_name} and \texttt{DESCRIPTION} defined in the previous step. The output is a list of 10 request form names and descriptions. We programmatically verify that the output is JSON parsable.

\item Eventually, for each request form dictionary (\texttt{FORM}) generated previously, we prompt llama3-70B to generate the \texttt{FunctionSchema} like this:
\begin{verbatim}
messages = [
  {"role": "system", "content": SYSTEM_MESSAGE},
  {"role": "user", "content": f"""
    [ENTERPRISE_NAME] is '{fake_name}'.
    {DESCRIPTION}
    Show me the FunctionSchema of the following REQUEST_FORM:
    ```json
    {FORM}
    ```
    First, rephrase the description to be more specific. The description must include
    rules, facts, and policies at {fake_name} about {FORM['name']}, such as when
    to fill this request, who will process the request, etc... The description must
    also include information about the parameters to help the user fill the form.
    For multiple choice parameters (enum), the description must explain the
    difference between each possible value.

    Once the description is written, remove all colon ( : ) characters from it.

    Second, write the yaml format of this FunctionSchema by replacing the description
    with your detailed version. Parameters cannot be nested objects, but return values
    can. The FunctionSchema must contain both required and optional parameters. 
    Write the output in yaml format. The output must be parsable as a FunctionSchema.
  """},
]
\end{verbatim}
with \texttt{SYSTEM\_MESSAGE}, \texttt{fake\_name}, \texttt{DESCRIPTION}, and \texttt{FORM} defined in the previous steps. We verify programmatically that the output is parsable as a \texttt{FunctionSchema}. An example of generated \texttt{FunctionSchema} can be seen in Step [5] of the Tapes in Figures~\ref{fig:teacher_tape} \& \ref{fig:student_tape} (Appendix~\ref{appendix:ghreat_tapes}).
\end{enumerate}

\subsection{List of Form-Filler Agent Thoughts and Actions \label{appendix:ghreatthoughts}}

\paragraph{Form-filler thoughts.}Below, we list the available thoughts for our Form-Filler TapeAgent and explain when they should be used:
\begin{itemize}
    \item \texttt{AnswerFromFunctionSchemaThought}: This thought premeditates answering the user's question that the agent knows how to answer based on the form description. Attributes:
        \begin{itemize}
            \item function: The name of the function to answer a question about.
        \end{itemize}
    \item \texttt{NoAnswerInFunctionSchemaThought}: This thought premeditates informing the user 
        that the agent does not know the answer to their question. Attributes:
        \begin{itemize}
            \item function: The name of the function.
        \end{itemize}
    \item \texttt{RefuseInexistentFunctionThought}: A thought that indicates that the query could not be resolved to a function.
    \item \texttt{RefuseInvalidFunctionParameterValueThought}: This thought premeditates 
        informing the user that a function parameter value is invalid.
        Attributes:
        \begin{itemize}
            \item function: The name of the function.
            \item parameter: The name of the parameter the user tried to set a value for.
            \item parameter\_value: The value the user tried to set for the parameter.
        \end{itemize}
    \item \texttt{RefuseSkippingParameterThought}: This thought premeditates informing the user
        that a function parameter cannot be skipped because it is required.
        Attributes:
        \begin{itemize}
            \item function: The name of the function.
            \item parameter: The name of the required parameter that the user tried to skip.
        \end{itemize}
    \item \texttt{GatherValuesThought}: This thought records extracted parameters from the user's message and is mainly used in the multi-node Teacher agent.
    Attributes:
    \begin{itemize}
        \item function: The name of the function.
        \item parameters: Dictionary mapping parameter names to their extracted values.
    \end{itemize}
    \item \texttt{VerifyValuesThought}: This thought records the validity status of parameter values extracted from the user's message and is mainly used in the multi-node Teacher agent.
    Attributes:
    \begin{itemize}
        \item function: The name of the function.
        \item parameters: Dictionary mapping parameter names to their value, their validity, and the explanation when invalid.
    \end{itemize}
    \item \texttt{UpdateFunctionParametersThought}: This thought updates the values of the parameters of a function based only on the user's last message. It is used when the user provides NEW information about the parameters of a function or when the user wants to skip an optional parameter. Attributes:
        \begin{itemize}
            \item function: The name of the function.
            \item assign: The dictionary assignment of parameter names to their NEW values (optional, default: \{\}).
                If no NEW values are provided, set `assign' to \{\}.
            \item skip: The list of NEW optional parameter names to skip (optional, default: []).
                If no NEW parameters are skipped, set `skip' to [].
        \end{itemize}
        The `assign' attribute sets or updates parameter values based on the last user message. It is only applied to parameters that are newly filled or have updated values. Parameters that remain unchanged will not have the `assign' attribute.
        The `skip' attribute is set for optional parameters when the user wants to ignore or skip an optional parameter. It is only applied to new optional parameters that the user wants to skip and will not be used for parameters that were already skipped.
    \item \texttt{RequestFunctionThought}: This thought premeditates requesting the user to select a function. Attributes:
    \begin{itemize}
        \item functions: The list of available functions the user can select.
    \end{itemize}
    \item \texttt{RequestFunctionParametersThought}: This thought premeditates requesting the user to provide a value for one parameter of a given function.
        Attributes:
        \begin{itemize}
            \item function: The name of the function to request parameters for.
            \item parameters: A list containing only one value which is the parameter to request.
        \end{itemize}
    \item \texttt{RequestFunctionCallConfirmationThought}: This thought is a preliminary step before calling a function.
        This step is ONLY used once ALL parameters are filled or skipped.
        Attributes:
        \begin{itemize}
            \item function: The name of the function.
        \end{itemize}
    \item \texttt{RequestExitConfirmationThought}: This thought is a preliminary step before exiting the dialogue. The message MUST be short and concise.
        The message should contain clear and explicit confirmation of the acceptance of the last value provided by the user.
\end{itemize}

\paragraph{Form-filler actions.} Here are the available actions our Form-Filler TapeAgent can take.
\begin{itemize}
    \item \texttt{ResolveFunctionAction}:       An action that resolves which function candidates are compatible with a given user query.
      Attributes:
      \begin{itemize}
          \item query: The query to resolve the function with, by default the user's last message.
          \item result: A list of candidate functions that are compatible with the query.
      \end{itemize}
    \item \texttt{InspectFunctionAction} An action that inspects the schema of a function given its name.
    This action inspects and returns the schema of the function,
    which contains its description and the names and types of the function's parameters.
    Attributes:
    \begin{itemize}
        \item function: The name of the function.
        \item result: The schema of the function.
    \end{itemize}
    \item \texttt{PromptUserForTextMessageAction}: An action that asks the user to enter a message in a text input field.
        This action is used to capture free-form text input from the user through a
        standard text input field. This action must be used in every single agent's response.
        Attributes:
        \begin{itemize}
            \item prompt: The message from the agent to the user.
            \item result: User message
        \end{itemize}
    \item \texttt{CallFunctionAction}:     This action calls a function with all available parameter values.
    Attributes:
    \begin{itemize}
        \item function: The name of the function to call.
        \item result: Result of the function call
    \end{itemize}
    \item \texttt{ExitAction}: With this action the agent indicates that the dialogue has ended.
        No further steps will be executed after this step.
        Attributes:
        \begin{itemize}
            \item text: The message from the agent to inform the user that the dialogue has ended.
        \end{itemize}
\end{itemize}

\newpage

\subsection{User Agents}
\label{appendix:user_behaviors}

\begin{table}[h]
\caption{The list of user agents implemented for all domains. The User LLM is a prompted Llama-3.1-405B-Instruct model.}
\label{tab:user_behaviors}
\centering
\begin{tabular}{ll}
\toprule
\textbf{Agent Name} & \textbf{Behavior} \\ \hline
UserInitMessageAmazing & \begin{tabular}[c]{@{}l@{}}At the begining of the conversation, the user should\\ request a specific intent and provide values for some\\ of the parameters in that request.\end{tabular}   \\ \hline
UserInitMessageShort & \begin{tabular}[c]{@{}l@{}}At the beginning of the conversation, the user should\\ request a specific intent in a short message.\end{tabular}   \\ \hline
UserInitMessageAsk & \begin{tabular}[c]{@{}l@{}}At the beginning of the conversation, the user should\\ ask the agent what it can do.\end{tabular}   \\ \hline
UserBadInitMessage & \begin{tabular}[c]{@{}l@{}}At the beginning of the conversation, the user should\\ ask to do something impossible.\end{tabular}   \\ \hline
UserHappyPath & The user correctly answers the agent question.   \\ \hline
UserMultislotInstruct & \begin{tabular}[c]{@{}l@{}}The user replies to the agent question and provides\\ values for additional parameters in the request.\end{tabular}   \\ \hline
UserMultislotFuture & \begin{tabular}[c]{@{}l@{}}The user provides values for additional\\ parameters in the request.\end{tabular}   \\ \hline
UserChangesSlot & \begin{tabular}[c]{@{}l@{}}The user changes the value to a previously\\ set request parameter.\end{tabular}   \\ \hline
UserAdjectiveOrPosition & \begin{tabular}[c]{@{}l@{}}For categorical parameters, the user answers the agent\\ but instead of giving the choice name, uses an\\ adjective or a positional number that defines its choice.\end{tabular}   \\ \hline
UserSkipsOptional & For optional parameters, the user asks to skip it.   \\ \hline
UserAsksAboutDocs & \begin{tabular}[c]{@{}l@{}}The user does not answer the agent, and instead\\ asks a question that can be answered based\\ on the request documentation.\end{tabular}   \\ \hline
UserAsksAboutFilledParameters & \begin{tabular}[c]{@{}l@{}}The user does not answer the agent, and instead\\ asks a question about the parameters filled,\\ or still missing.\end{tabular}   \\ \hline
UserAsksForHelp & \begin{tabular}[c]{@{}l@{}}For categorical parameters, the user ask the\\ agent for help in choosing a value based on\\ its problem.\end{tabular}   \\ \hline
UserCorrectValueInvalidValue & \begin{tabular}[c]{@{}l@{}}The user answers the agent correctly, but also\\ gives an invalid value for a categorical\\ parameter not yet filled.\end{tabular}   \\ \hline
UserInvalidValue & \begin{tabular}[c]{@{}l@{}}For categorical parameters, the user gives\\ an impossible choice.\end{tabular}   \\ \hline
UserChangesTopic & \begin{tabular}[c]{@{}l@{}}The user does not answer the agent's\\ question and change the topic of\\ conversation instead.\end{tabular}   \\ \hline
UserSkipsUnskipable & For required parameters, the user asks to skip it.   \\ \hline
UserIncorrectDate & \begin{tabular}[c]{@{}l@{}}For date/time parameters, the user gives\\ an invalid date.\end{tabular}   \\ \hline
UserIncorrectWeb & \begin{tabular}[c]{@{}l@{}}For technical parameters such as\\ url/email, the user gives a invalid value.\end{tabular}   \\
\bottomrule
\end{tabular}
\end{table}

\subsection{Choosing the teacher model}
\label{appendix:teachers}

The Teacher agent uses a Llama-405B-Instruct model\footnote{\url{https://huggingface.co/meta-llama/Llama-3.1-405B-Instruct-FP8}} that is carefully prompted to perform the form-filling task according to all our metrics (Section \ref{subsec:ghreat_metrics}).
We experimented with various language models such as GPT4o, and Mistral-Large, but according to human evaluation, Llama 405B was on par or better than other models.
We explored two different prompting techniques. The first (referred to as `long-node' in Table~\ref{tab:teacher_bakeoff}) consists of crafting one long and detailed prompt explaining how the agent should make a thought plan and answer the user (single node TapeAgent). The other prompting technique explored (referred to as `multi-node' in Table~\ref{tab:teacher_bakeoff}) consists in decomposing the task into multiple smaller subtasks, each handled by a different node. The multi-node agent is first tasked to identify potential parameter values provided by the last user message, then it is prompted to verify whether the provided values are correct or not, eventually, the agent is prompted to note valid information, refuse invalid information, and ask the user for the next parameter to fill.
We generated agent messages as a continuation to $1016$ unfinished conversations where the last user message behaved according to specific behaviors and sent these messages to human labelers. Results show that the multi-node approach yields better performance overall, so we decided to keep this prompting strategy as our final Teacher agent in Section~\ref{subsec:ghreat_design}.

\begin{table}[h]
\caption{Evaluation of the Teacher agent with two prompting methods (long-node \& multi-node) across all user behaviors on 1016 test domain conversations.}
\label{tab:teacher_bakeoff}
\centering
\begin{tabular}{ll|cc}
\toprule
 &  & \multicolumn{2}{c}{\textbf{GREADTH metric}} \\
\textbf{Author of last user message} & \textbf{Count} & long-node & multi-node \\ \hline
UserInitMessageAmazing & 46 & \cellcolor[HTML]{ABDCB9}60.87\% & \cellcolor[HTML]{7BC890}86.96\% \\
UserInitMessageShort & 48 & \cellcolor[HTML]{7AC88F}87.50\% & \cellcolor[HTML]{77C68C}89.58\% \\
UserInitMessageAsk & 46 & \cellcolor[HTML]{73C589}91.30\% & \cellcolor[HTML]{7FCA94}84.78\% \\
UserBadInitMessage & 46 & \cellcolor[HTML]{67C07F}97.83\% & \cellcolor[HTML]{63BE7B}100.00\% \\
UserHappyPath & 186 & \cellcolor[HTML]{96D3A7}72.58\% & \cellcolor[HTML]{77C78D}89.25\% \\
UserMultislotInstruct & 48 & \cellcolor[HTML]{9DD6AD}68.75\% & \cellcolor[HTML]{82CB96}83.33\% \\
UserMultislotFuture & 46 & \cellcolor[HTML]{AFDDBD}58.70\% & \cellcolor[HTML]{8FD0A1}76.09\% \\
UserChangesSlot & 46 & \cellcolor[HTML]{97D4A8}71.74\% & \cellcolor[HTML]{83CB97}82.61\% \\
UserAdjectiveOrPosition & 46 & \cellcolor[HTML]{AFDDBD}58.70\% & \cellcolor[HTML]{8FD0A1}76.09\% \\
UserSkipsOptional & 46 & \cellcolor[HTML]{97D4A8}71.74\% & \cellcolor[HTML]{97D4A8}71.74\% \\
UserAsksAboutDocs & 46 & \cellcolor[HTML]{7FCA94}84.78\% & \cellcolor[HTML]{6FC386}93.48\% \\
UserAsksAboutFilledParameters & 48 & \cellcolor[HTML]{82CB96}83.33\% & \cellcolor[HTML]{86CC99}81.25\% \\
UserAsksForHelp & 45 & \cellcolor[HTML]{D2EBDB}40.00\% & \cellcolor[HTML]{CEE9D7}42.22\% \\
UserCorrectValueInvalidValue & 45 & \cellcolor[HTML]{A9DBB7}62.22\% & \cellcolor[HTML]{8CCF9F}77.78\% \\
UserInvalidValue & 46 & \cellcolor[HTML]{A7DAB6}63.04\% & \cellcolor[HTML]{73C589}91.30\% \\
UserChangesTopic & 46 & \cellcolor[HTML]{8FD0A1}76.09\% & \cellcolor[HTML]{7BC890}86.96\% \\
UserSkipsUnskipable & 48 & \cellcolor[HTML]{7AC88F}87.50\% & \cellcolor[HTML]{6BC282}95.83\% \\
UserIncorrectDate & 46 & \cellcolor[HTML]{C3E5CE}47.83\% & \cellcolor[HTML]{F7FAFB}19.57\% \\
UserIncorrectWeb & 42 & \cellcolor[HTML]{FCFCFF}16.67\% & \cellcolor[HTML]{EFF7F4}23.81\% \\ \hline
\textbf{ALL\_BEHAVIORS} & \textbf{1016} & \cellcolor[HTML]{9CD5AC}\textbf{69.39\%} & \cellcolor[HTML]{8BCE9D}\textbf{78.54\%} \\
\bottomrule
\end{tabular}
\end{table}

\newpage

\subsection{Teacher and Student Tapes}
\label{appendix:ghreat_tapes}

\end{document}